\documentclass[10pt,journal,compsoc]{IEEEtran}
\ifCLASSOPTIONcompsoc
  \usepackage[nocompress]{cite}
\else
  \usepackage{cite}
\fi
\ifCLASSINFOpdf
\else
\fi
\usepackage{amsmath,amsfonts}
\usepackage{algorithmic}
\usepackage{array}
\usepackage{textcomp}
\usepackage{stfloats}
\usepackage{url}
\usepackage{verbatim}
\usepackage{graphicx}
\usepackage{amssymb}
\usepackage{bm}
\usepackage{subfigure}
\usepackage{multirow}
\usepackage{color}
\usepackage{cite}
\usepackage{tabularx}
\usepackage{hyperref}

\DeclareMathOperator*{\argmin}{arg\,min}
\def\thetaset{\bm{\theta}}
\def\Thetaset{\bm{\Theta}}

\def\Xt{\widetilde{\mathbf{X}}}
\def\Yt{\widetilde{\mathbf{Y}}}
\def\xt{\widetilde{\mathbf{x}}}
\def\yt{\widetilde{\mathbf{y}}}
\def\xhat{\hat{\mathbf{x}}}
\def\yhat{\hat{\mathbf{y}}}

\def\Xhat{\hat{\mathbf{X}}}
\def\Yhat{\hat{\mathbf{Y}}}

\def\N{\mathcal{N}}
\def\NxX{\N_{\xt_i,K}^{\Xt_l}}
\def\NXX{\N_{\Xt_l,K}^{\Xt_l}}

\def\NXY{\N_{\Xt_l,K}^{\Yt_l}}
\def\NxXhat{\N_{\xhat_i,K}^{\Xhat_l}}

\def\p{\mathbf{p}}
\def\x{\mathbf{x}}
\def\y{\mathbf{y}}
\def\s{\mathbf{s}}
\def\e{\mathbf{e}}

\def\v{\mathbf{v}}
\def\X{\mathbf{X}}
\def\Y{\mathbf{Y}}
\def\I{\mathbf{I}}
\def\E{\mathbf{E}}
\def\S{\mathbf{S}}
\def\R{\mathbb{R}}
\def\V{\mathbf{V}}
\def\Tr{\mathrm{Tr}}
\newcommand{\znote}{\textcolor{black}}

\begin{document}

\title{TCDM: Transformational Complexity Based Distortion Metric for Perceptual Point Cloud Quality Assessment}

%
%

\author{Yujie Zhang, Qi Yang, Yifei Zhou, Xiaozhong Xu, Le Yang, Yiling Xu, ~\IEEEmembership{Member,~IEEE}

\thanks{Y. Zhang, Y. Xu are with the Cooperative Medianet Innovation Center, Shanghai Jiao Tong University, Shanghai, China, (e-mail: yujie19981026@sjtu.edu.cn, yl.xu@sjtu.edu.cn)}
\thanks{Q. Yang is with Tencent MediaLab, Shanghai, China, (email: chinoyang@tencent.com) }
\thanks{Y. Zhou is with Shanghai Maritime Univeristy, Shanghai, China, (email: 202110310163@stu.shmtu.edu.cn) }
\thanks{X. Xu is with Tencent MediaLab, Palo Alto, America, (email: xiaozhongxu@tencent.com) }
\thanks{L. Yang is with the Department of Electrical and Computer Engineering, University of Canterbury, Christchurch, New Zealand, (email: le.yang@canterbury.ac.nz)}
}

%
%

\markboth{Journal of \LaTeX\ Class Files,~Vol.~14, No.~8, August~2015}%
{Shell \MakeLowercase{\textit{et al.}}: Bare Demo of IEEEtran.cls for Computer Society Journals}
%



\IEEEtitleabstractindextext{%
\begin{abstract}

The goal of objective point cloud quality assessment (PCQA) research is to develop quantitative metrics that measure point cloud quality in a perceptually consistent manner. Merging the research of cognitive science and intuition of the human visual system (HVS), in this paper, we evaluate the point cloud quality by measuring the complexity of transforming the distorted point cloud back to its reference, which in practice can be approximated by the code length of one point cloud when the other is given. For this purpose, we first make space segmentation for the reference and distorted point clouds based on a 3D Voronoi diagram to obtain a series of local patch pairs. Next, inspired by the predictive coding theory, we utilize a space-aware vector autoregressive (SA-VAR) model to encode the geometry and color channels of each reference patch with and without the distorted patch, respectively. Assuming that the residual errors follow the multi-variate Gaussian distributions, the self-complexity of the reference and transformational complexity between the reference and distorted samples are computed using covariance matrices. Additionally, the prediction terms generated by SA-VAR are introduced as one auxiliary feature to promote the final quality prediction. The effectiveness of the proposed transformational complexity based distortion metric (TCDM) is evaluated through extensive experiments conducted on five public point cloud quality assessment databases. The results demonstrate that TCDM achieves state-of-the-art (SOTA) performance, and further analysis confirms its robustness in various scenarios. The code will be publicly available at \url{https://github.com/zyj1318053/TCDM}.

\end{abstract}

\begin{IEEEkeywords}
Point cloud quality assessment, image quality assessment, predictive coding, vector autoregressive model
\end{IEEEkeywords}}

\maketitle

\IEEEdisplaynontitleabstractindextext

%
\IEEEpeerreviewmaketitle


\section{Introduction}\label{sec:intro}

%
%
%
%
Three-dimensional (3D) point cloud data has attracted much attention in recent years due to its wide applications, such as autopilot, virtual reality (VR), and 3D free-viewpoint broadcasting. In practical application scenarios, point cloud data is inevitably subject to a variety of distortions during acquisition, compression, transmission, and rendering stages, which usually impairs the perceptual quality of the human visual system (HVS) \cite{shan2022gpa}. To establish the guidance of multiple point cloud applications targeted for human vision, point cloud quality assessment (PCQA)
has been widely explored, which can be further categorized into subjective PCQA and objective PCQA. Although subjective PCQA can provide accurate predictions, its application is limited due to the high cost in terms of time and money. In comparison, objective PCQA intends to design automatic approaches called PCQA metrics that can predict point cloud quality consistently with subjective evaluation, which is more convenient in practice due to lower cost and higher speed. PCQA metrics play a vital role in quality of experience (QoE) oriented data processing. For instance, for point cloud compression or transmission, PCQA metrics can be used to derive constraints to achieve a better trade-off between quality and compression rate or transmission bandwidth \cite{liu2020model, liu2022no,su2023bitstream}. Moreover, PCQA metrics can also serve as loss functions to guide model optimization in learning-based point cloud generation tasks, such as spatial upsampling and completion \cite{yang2021mped,fan2017point,wu2021density}. Therefore, it is highly desired to develop reliable and effective objective PCQA metrics correlated with human perception. In this paper, we focus on designing a full-reference (FR) PCQA metric for cases where the high-quality original point cloud is available as the reference for evaluating the quality of its distorted versions with the same poses.

\begin{figure}
    \centering
    \subfigure[]{\includegraphics[width=1\linewidth]{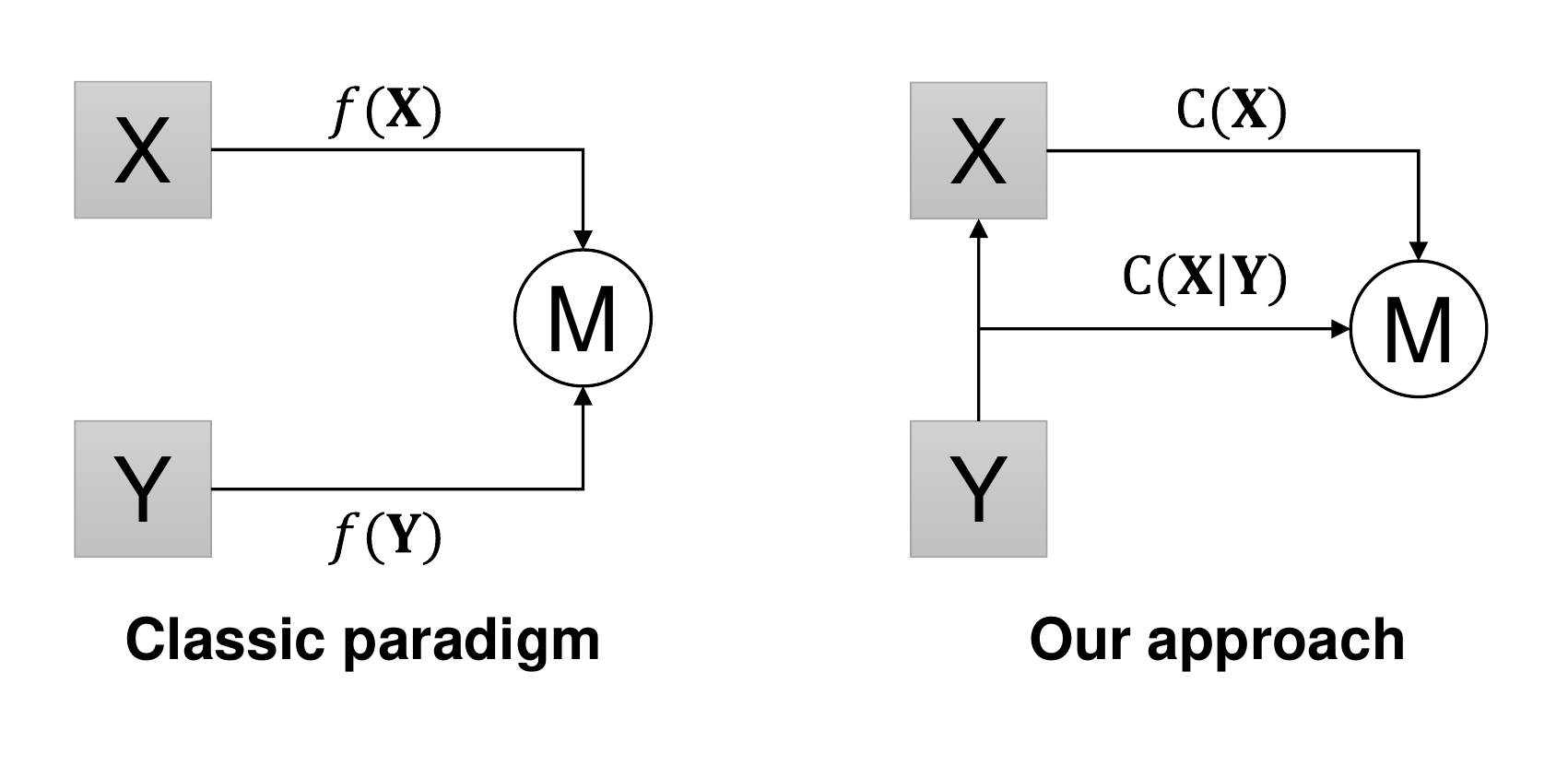}}
    \subfigure[]{\includegraphics[width=1\linewidth]{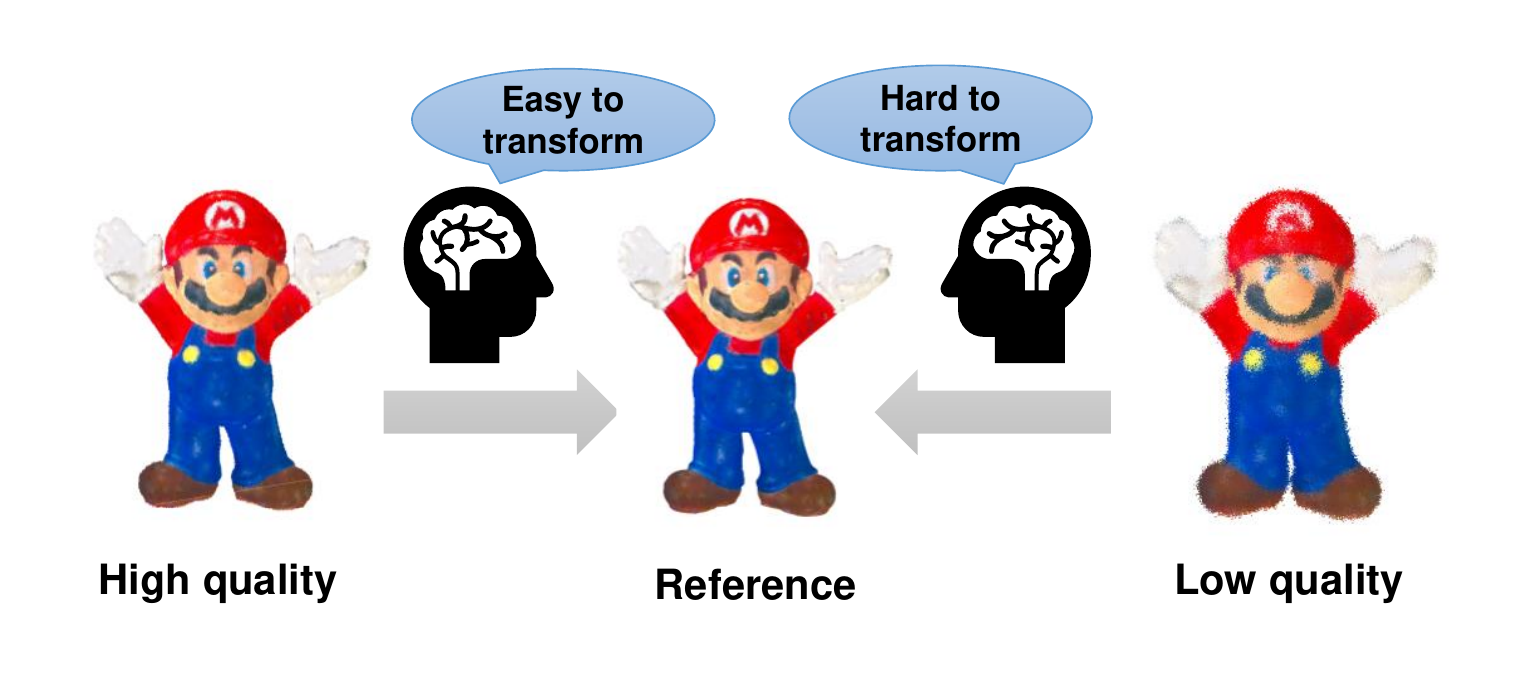}}
    \caption{We consider the FR-PCQA problem from the perspective of transformational complexity. The greater the effort required to transform the distorted point cloud to its reference, the lower the quality of the distorted sample. (a) The paradigms of previous FR-PCQA research and our approach;
    (b) illustration of the motivation considered in this paper.}
    \label{fig:brain_model}
\end{figure}

\znote{Early FR-PCQA metrics, e.g., point-to-point (p2po) and $\rm PSNR_{YUV}$ \cite{MPEGSoft}, try to quantify point cloud distortions by calculating the geometry or color difference between each point of the distorted/reference point cloud and its nearest point in the reference/distorted counterpart, which results in a symmetric prediction.} However, these simple point-based metrics do not correlate well with human perception \cite{yang2020inferring}. To solve this problem, some new PCQA metrics have been proposed, which can generally be categorized into 2D-based metrics \cite{lavoue2015efficiency, de2017motion,alexiou2019exploiting,javaheri2022joint,liu2022perceptual,yang2020predicting,he2021towards, Yang_2022_CVPR, liu2021pqa}, and 3D-based metrics \cite{meynet2019pc,meynet2020pcqm,yang2020inferring,zhang2021ms,diniz2020local,9123076,alexiou2021pointpca} according to whether the point cloud is projected into images. The 2D-based metrics first project the point cloud onto 2D planes at different viewing angles and then take advantage of the well-developed image processing tools to assist in PCQA.
As for the 3D-based metrics, they try to predict point cloud quality by modeling spatial properties. Due to the irregularity of point cloud data, 3D-based metrics normally need to construct efficient data structures (e.g., surface \cite{meynet2020pcqm}, graph \cite{yang2020inferring}, mesh \cite{lavoue2011multiscale, nehme2020visual}) to facilitate subsequent processing.

In general, the paradigm of most existing PCQA metrics, whether they are 2D- or 3D-based, is to derive the quality score by comparing some perceptually sensitive features (e.g., gradient, curvature) extracted \textit{separately} from the reference and distorted point clouds. As shown on the left side of Fig. 1 (a), given two point clouds $\X$ and $\Y$, the above paradigm can formulate their difference $d(\X,\Y)$ as $d(\X,\Y)=M(f(\X),f(\Y))$, where $f(\cdot)$ denotes the feature extraction operator and $M(\cdot)$ denotes the pooling operator. 
The success of such methods is highly dependent on the effectiveness of the extracted features relying on $f(\cdot)$. However, due to the irregular 3D structure of point cloud and the abundance of distortion types, a single feature is usually insufficient to handle various types of distortion. To improve the robustness and accuracy of the metric, there is a tendency to use as many features as possible (e.g., PCQM \cite{meynet2020pcqm} contains 8 features and PointPCA \cite{alexiou2021pointpca} contains 32 features), which can cause the metrics to become increasingly bloated.

\subsection{Motivation}
In this paper, we propose to solve the problem of PCQA from a new perspective, namely, transformational complexity. In cognitive science, recent studies indicate that the similarity between two entities can be quantified using the "\textit{complexity}" required to transform the representation of one into that of the other \cite{hahn2003similarity}, \cite{goldstone2012similarity}, which is called "\textit{transformational complexity}" in this paper. According to prior empirical research, the more difficult the transformation from one entity to the other is, the less similar they would be. For instance, Hahn et al. \cite{hahn2003similarity} found that the transformational complexity accurately predicted similarity ratings between simple graphics in three experiments. For more complicated scenarios such as image processing, many similarity metrics based on the transformational complexity also show competitive performance in several applications, including image retrieval \cite{guha2014image}, classification \cite{lan2005image}, etc.

Inspired by the success of the transformational complexity in the above scenarios, in this paper, we intend to investigate its utility in PCQA. Intuitively, we suppose that the quality of one distorted point cloud can be quantified as the complexity or the amount of effort of transforming it into its corresponding reference. As shown in Fig. \ref{fig:brain_model} (b), denoising a low-quality sample disturbed by severe geometry Gaussian noise (GGN) is obviously more difficult than denoising a high-quality sample disturbed by slight noise because severe geometry noise
damages the structure of some regions (e.g., eyes), which brings higher complexity for recovery. More detailed visual examples can be found in Fig. \ref{fig:prediction_term}. According to Fig. \ref{fig:prediction_term}, given one reference point cloud patch, we intend to reconstruct it from multiple distorted patches impaired by GGN at three levels. As the level of distortion increases, it becomes evident that the reconstructed results (called cross-prediction terms) of the distorted patches exhibit greater disparity from the reference patch. We consider that the worse reconstruction results (i.e., with larger residuals) indicate larger transformational complexity, which in turn represents lower visual quality.  To reduce the influence of the intrinsic characteristics of different samples, the \textit{self-complexity} of the reference can be introduced as a normalization factor. Therefore, our method can be formulated as $d'(\X,\Y) = M(C(\X),C(\X|\Y))$ shown on the right side of Fig. \ref{fig:brain_model} (a), where $C(\X)$ and $C(\X|\Y)$ are the self-complexity of $\X$ and the transformational complexity between $\X$ and $\Y$. Compared to the classic FR-PCQA paradigm, our approach avoids the complicated process of feature selection (that is, finding $f(\cdot)$) and instead regards the quality evaluation problem as inferring the difficulty of recovering the reference from its distorted version.

One critical problem in our work is how to calculate the transformational complexity between two point clouds. The basis of transformational complexity can be found in a particular branch of mathematics known as Kolmogorov complexity theory \cite{li2008introduction}. However, Kolmogorov complexity is in general an intractable quantity \cite{vitanyi2020incomputable}. In practice, it is often approximated by the code length of compressed data \cite{li2004similarity}. Intuitively, the more a given data can be compressed, the lower its complexity would be. Therefore, the transformational complexity between two entities is usually described as the code length of one entity when the other is given. For instance, Guha et al. \cite{guha2014image} approximated the sparsity of an image using the overcomplete dictionary extracted from the other image, where the sparse coefficients are used as a measure of the transformational complexity. One important advantage of these compression-based approaches is their correlation with cognitive mechanisms. Recent developments in brain theory and neuroscience \cite{spratling2017review}, \cite{friston2009predictive} show that human perception of external stimuli can be modeled as a process of predictive coding, which can be accounted for by the principles of parsimony and redundancy reduction. Hence, deriving the transformational complexity via compression-based approaches possesses a tight connection to human perception. In our metric, we utilize the predictive coding theory to compute the transformational complexity. 

\subsection{Approach} We propose a novel FR-PCQA metric named the transformational complexity based distortion metric (TCDM). Our metric is
employed by the following three steps:

First, considering that local processing is more effective for quality assessment, we segment the reference and distorted point clouds into multiple local patch pairs based on one 3D Voronoi diagram (VD), which can ensure non-overlapping divisions and avoid the omission of points. 

Secondly, we utilize an SA-VAR model to encode the XYZ and RGB channels of each reference patch in cases with and without the corresponding distorted patch. Specifically, for each point in one reference patch, we use its neighbors in the reference and those in the distorted patch to predict itself using one autoregressive (AR) model, respectively.

Finally, two complexity terms, i.e., the self-complexity of the reference and the transformational complexity between the reference and distorted samples, are calculated to derive a complexity-based similarity metric. The prediction terms generated by SA-VAR are introduced as an auxiliary feature to achieve the final quality index. 

Experimental results on five publicly available PCQA databases confirm that the proposed metric achieves state-of-the-art performance. 

\subsection{Contributions}
The main contributions of the paper are summarized as follows.

$\bullet$ We propose a novel point cloud quality metric called TCDM. Compared to existing PCQA metrics, our metric formulates the problem of quality evaluation from the perspective of measuring the transformational complexity, which is more effective than these methods that rely on separately extracted features.

$\bullet$ To compute the transformational complexity, we propose an SA-VAR model to encode both the geometry and the color information of point cloud data. We introduce a predefined spatial weight to strengthen the awareness for irregular structure, and we utilize a multi-variate vector to capture the correlation between multiple channels. 

$\bullet$ Our metric shows reliable performance in five publicly accessible databases. Further analyses reveal the model's generalization capability under various parameter settings.

The remainder of this paper proceeds as follows. Section \ref{sec:related work} presents the related work. Section \ref{sec:problem formulation} formulates the FR-PCQA problem and the core idea of our method.  Section \ref{sec:the proposed method} introduces the implementation details of the proposed method. Section \ref{sec:experiment results} gives the experiment results. Section \ref{sec:conclusion} draws the conclusion.

\begin{figure*}
    \centering
    \includegraphics[width=1\linewidth]{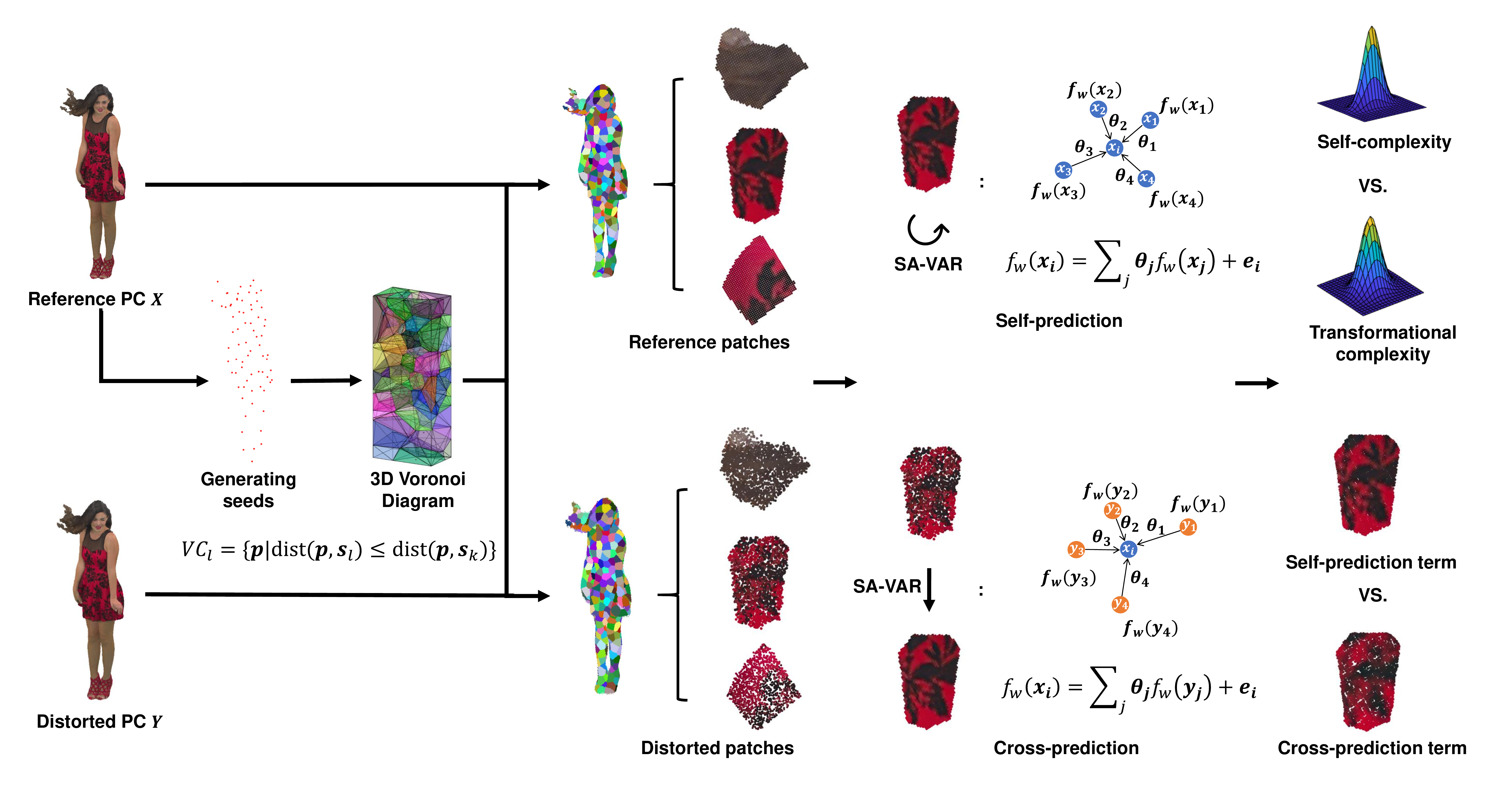}
    \caption{The framework of the proposed method. A point set is first sampled from the reference to serve as the generating seeds of the 3D Voronoi diagram. Utilizing the Voronoi diagram, both the reference and distorted samples can be segmented into a series of local patch pairs. For each local patch pair, an SA-VAR model is employed to perform the self-prediction for the reference patch and the cross-prediction between the patch pair. Two complexity terms and the prediction terms generated by the SA-VAR are leveraged to derive the local quality indices. All local quality indices are then fused to obtain the final quality score.}\label{fig:framework}
\end{figure*}

\section{Related Work}\label{sec:related work}
A handful of metrics have been proposed for FR-PCQA, which can be roughly categorized as 2D-based metrics and 3D-based metrics. Note that we classify early point-wise metrics (e.g., p2po and $\rm{PSNR}_{YUV}$) into 3D-based metrics because they do not have a point cloud projection operation. We give a brief review of these metrics in the following.

\textbf{2D-based metrics:}
2D-based metrics rely on quantifying point cloud distortion using projected images. To simulate the multi-view nature of observing 3D objects, researchers usually project the point cloud onto 2D planes at different viewing angles. In \cite{de2017motion}, Queiroz et al. projected the reference and distorted point clouds onto the six faces of a bounding box enclosing the point clouds and then concatenated the corresponding projected images. 2D PSNR was used to measure the distortions between the corresponding distorted and reference projected images. In \cite{alexiou2019exploiting}, Alexiou et al. conducted a study on camera layouts around the point cloud, exploiting the impact of the number of projected images in 2D-based metrics. Different IQA methods (e.g., PSNR, SSIM \cite{wang2004image} and VIFP \cite{sheikh2006image}) were utilized to predict the quality of projected images and the results showed that a larger number of projected views does not necessarily lead to better predictions of point cloud quality. To eliminate the misalignment between the reference and distorted projected images, Javaheri et al. \cite{javaheri2022joint} proposed to assign the same geometry to both the reference and distorted point clouds before the projection operation. More recently, to distinguish the background region and point cloud region in projected images, Liu et al. \cite{liu2022perceptual} proposed one new metric based on the principle of information content weighted structural similarity (IW-SSIM) \cite{wang2010information}. 

Apart from the most commonly used color texture projected images, some geometry-related maps, such as depth and curvature maps, are also introduced into some 2D-based metrics to achieve more robust quality prediction. In \cite{yang2020predicting},
Yang et al. chose to project the 3D point cloud onto six perpendicular image planes of a cube to obtain both color texture and depth maps. Curvature maps used as an auxiliary term in \cite{he2021towards}. By introducing additional information to aid in quality assessment, the above metrics usually outperformed traditional IQA metrics that only use color texture projected images. 

\textbf{3D-based metrics:}
3D-based metrics predict point cloud quality by modeling its 3D properties. Point-wise metrics were first applied to evaluate point cloud distortions, including p2po, point-to-plane (p2pl) and $\rm PSNR_{YUV}$ \cite{MPEGSoft}. Specifically, p2po and $\rm PSNR_{YUV}$ calculated the distance between the geometric coordinates or YUV values of each point pair. In comparison, p2pl computed the geometric distance between two points along the normal direction of the point cloud surface, resulting in smaller errors for points closer to the surface. Researchers have proposed some improvements for the above metrics in recent years. For example, to alleviate the influence of outlier points, Javaheri et al. \cite{javaheri2020generalized} proposed to use the Hausdorff distance for a specific percentage of data rather than the entire data set. In \cite{wang2023improving}, Wang et al. incorporated just noticeable difference theory (JND) to derive the noticeable possibility of each point, leading to a better correlation with subjective judgement. 

\znote{Point-wise metrics are not well correlated with subjective ratings, leading to unreliable prediction performance. Thus, metrics that involve structural features have been developed to better predict subjective perception. In \cite{alexiou2018point}, Alexious proposed a new metric named plane-to-plane (pl2pl), which utilized the normal vector of two points to derive the angular similarity.}  Meynet et al. proposed PC-MSDM \cite{meynet2019pc} based on the local curvature similarity. Utilizing PC-MSDM, Meynet et al. extended it to PCQM \cite{meynet2020pcqm}, which used three curvature-related features and five color-related features to improve the prediction robustness. By constructing local graphs based on geometry information and regarding color values as graph signals, Yang et al. \cite{yang2020inferring} developed GraphSIM to model the overall perception of HVS. A multi-scale version of GraphSIM was proposed by Zhang et al. in \cite{zhang2021ms}, which further took into account the multi-scale characteristics of human perception. In \cite{9123076}, Diniz et al. adapted the local binary pattern (LBP) to point clouds. A variant descriptor called local luminance patterns (LLP) was proposed in \cite{diniz2020local}, introducing a voxelization step. PointPCA \cite{alexiou2021pointpca} proposed the use of geometric descriptors based on the principal component analysis (PCA) to estimate structural distortions in point cloud contents. 

As suggested in Section \ref{sec:intro}, most of the above metrics
focus on how to extract more effective features. In other words, they intend to map the reference and its distorted version separately into two representations located in the same cognitive space that correlates well with human perception, and then represent quality in terms of the distance between the above two representations. In comparison, in this paper, we approach the FR-PCQA problem from a new perspective, i.e.,
evaluating the transformational complexity of obtaining the reference from its distorted version.

\section{Problem Formulation}\label{sec:problem formulation}

In this section, we illustrate some basic concepts. We first present the formulation of the FR-PCQA problem; then we detail the concept of the transformational complexity and the predictive coding theory and employ them in the FR-PCQA problem.

\subsection{Distortion Quantification of 3D Point Clouds}\label{sec:problem_formulation_1}
{\bf Point Cloud Representation. }Let $\X$ be a 3D point cloud with $N$ points: $\X = \{\x_1, \cdots, \x_N\} \in \R^{N\times6}$, where each $\x_i \in \R^6$ is a vector with 3D coordinates and three-channel color attributes, therefore, $\x_i=[x, y, z, R, G, B]\equiv[\x_i^O, \x_i^I]$, where $\x_i^O=[x, y, z]$ and $\x_i^I=[R, G, B]$. The superscript “O” stands for geometric \textit{occupancy}, and “I” stands for color \textit{intensity}.

{\bf Full-Reference Point Cloud Quality Assessment. }
Given one reference point cloud $\X$ and its distorted version $\Y$, an objective distortion quantification aims to measure the difference between the two samples. Considering that point clouds used for human perception tasks target visualization, FR-PCQA intends to evaluate the visual quality of the distorted $\Y$ with respect to its reference $\X$ . Mathematically, the purpose of FR-PCQA is to find a measure $d(\cdot)$ that satisfies

\begin{equation}\label{eq:hv_formulation}
  \phi(d(\X,\Y))=s,
\end{equation}
where $s$ denotes the ground truth data called mean opinion score (MOS) acquired from subjective experiments; $\phi(\cdot)$ denotes the nonlinear mapping to be learned, such as logistic regression or cubic regression.

As mentioned in Sections \ref{sec:intro} and \ref{sec:related work}, most existing FR-PCQA metrics try to extract perceptually effective features separately from the reference and distorted point cloud. Accordingly, we have 

\begin{equation}
    d(\X,\Y) = M(f(\X),f(\Y)).
\end{equation}
In practice, $f(\cdot)$ intends to extract those visually sensitive features, such as gradient and curvature, and $M(\cdot)$ denotes the pooling operation, such as the Minkowski pooling or similarity pooling \cite{wang2004image}.

\subsection{Transformational Complexity}\label{sec:trans_complexity_theory}
The complexity of an object is related to its randomness or redundancy. For example, the binary string 110101 is considered more complex than the string 010101, because the latter consists of repeating patterns and is consequently less random. The Kolmogorov complexity formalizes this concept. Specifically, given a finite signal $\x$, its Kolmogorov complexity $K(\x)$ is defined as the length of the shortest program complexity that can effectively be produced on a universal computer, such as a Turing machine \cite{bennett1998information}. Based on the above definition,
given two signals $\x$ and $\y$, the transformational complexity between them is developed using the conditional Kolmogorov complexity $K(\x|\y)$, which is defined as the length of the shortest program in a universal computer program to generate $\x$ when $\y$ is known.

Due to the non-computable nature of the Kolmogorov complexity \cite{vitanyi2020incomputable}, compression-based methods usually serve as an alternative to compute the transformational complexity. One pioneering work is the normalized compression distance (NCD) \cite{li2004similarity}, which considers the transformational complexity $C(\x|\y)$ as the code length of $\x$ when $\y$ is given. Specifically, the NCD can be formulated as
\begin{equation}
    NCD(\x,\y) = \frac{\max{\{C(\x|\y),C(\y|\x)\}}}{\max{\{C(\x),C(\y)\}}},\nonumber
\end{equation}
where $C(\cdot)$ is a compressor and the denominator $\max{\{C(\x),C(\y)\}}$ serves as an normalized factor. $C(\x|\y)$ is defined as the compressed length of $\x$ when $\y$ is given.

The NCD metric is regarded as an effective \textit{similarity metric} and shows reliable performance in multiple applications such as clustering languages and music \cite{cilibrasi2005clustering}. Inspired by its success in one-dimensional (1D) data, various compression-based methods have been proposed for higher-dimensional data types such as image and video \cite{guha2014image}, \cite{nikvand2013image}. The key idea of these methods is how to encode the redundancy inherent in the data. In addition to the transformational complexity $C(\x|\y)$, the self-complexity of $\x$, i.e., $C(\x)$, is usually utilized as a normalization factor. In general, we can formulate these compression-based similarity measures as follows,
\begin{equation}\label{eq:compression_distance}
    d'(\x,\y)=M(C(\x),C(\x|\y)),
\end{equation}
where $d'(\x,\y)$ denotes the distance or similarity between two entities $\x$ and $\y$.

Although the transformational complexity has achieved great success for 1D and 2D signals, there is no similar research related to 3D data. Moreover, when we focus our attention on the PCQA task, it is also desirable to establish the connection between compression-based methods and visual perception. To solve the above problems, our work tries to apply the transformational complexity for PCQA task based on the predictive coding theory.

\subsection{Predictive Coding Theory}\label{sec:predictive_coding_theory}
Predictive coding is a leading theory on how the brain performs probabilistic inference \cite{spratling2017review}. It postulates that the human brain learns the statistical regularities inherent in the natural world and reduces redundancy by only encoding what is not predictable (that is, the residual errors in prediction) \cite{huang2011predictive}. The underlying assumption of this theory is that our cognitive process is governed by an internal generative model in the brain. With this internal model, the brain can generate the corresponding predictions for the visual scenes encountered. Mathematically, given an internal brain model for visual perception, denoted $G$, it can adjust its parameter $\Thetaset$ to minimize the negative logarithmic likelihood function $-\log{p(\x|\Thetaset,G)}$, which is usually described as “\textit{code length}” of $\x$ \cite{huang2011predictive} due to the same formulation as Shannon's entropy. For some predictive coding algorithms such as traditional linear predictive coding \cite{o1988linear}, \cite{1162641}, the optimal parameter $\hat{\Thetaset}$ can be solved by

\begin{equation}\label{eq:optimal_theta}
     \hat{\Thetaset} = \argmin_{\Thetaset}{-\log{p(\x|\Thetaset,G)}}.
\end{equation}

In general, the predictive coding theory shares similar principles of redundancy reduction with information compression, inspiring a large number of compression algorithms \cite{wang2010spatial}, \cite{schwarz2018emerging}. Therefore, we can model our visual perception as one compression process and relate the minimized negative log-likelihood function to the concept of complexity. To achieve the goal of PCQA, we need an efficient quality-related internal model $G$. In our work, the SA-VAR model is employed as $G$, which is described in detail in Section \ref{sec:the proposed method}. Then, given two point clouds $\X$ and $\Y$, we can derive the self-complexity of $\X$, $C(\X)$, via a self-prediction process, and the transformational complexity between two point clouds, $C(\X|\Y)$, via a cross-prediction process as follows,
\begin{equation}
\begin{aligned}
    C(\X) &=\min\nolimits_{\Thetaset_s}{-\log{p(\X|\Thetaset_s,G)}},\\
    C(\X|\Y) &=\min\nolimits_{\Thetaset_t}{-\log{p(\X|\Y,\Thetaset_t,G)}},
\end{aligned}
\end{equation}
where $\Thetaset_s$ and $\Thetaset_t$ denote the model parameters in the self-prediction and cross-prediction processes. We can derive one complexity-based similarity metric according to Eq. \eqref{eq:compression_distance}, i.e., $d'(\X,\Y)=M(C(\X),C(\X|\Y))$, which can serve as one quality representation for PCQA.

\section{Proposed Method}\label{sec:the proposed method}
In this part, we first introduce the overall framework of the proposed model in Section \ref{sec:overview}. Then, we present the implementation details for each module.

\vspace{-1cm}

\subsection{Architecture} \label{sec:overview}
We show the overall framework of our model in Fig. \ref{fig:framework}. Taking the
reference $\X$ and distorted point cloud $\Y$ as input, we first extract a sequence of points from $\X$ by the farthest point sampling (FPS)  \cite{eldar1997farthest}. Utilizing the sampling points as the generating seeds, we make the space segmentation to derive one 3D Voronoi diagram (VD), which consists of many Voronoi cells (VCs). Each cell in the VD bounds one local patch pair derived from $\X$ and $\Y$. Next, for each local patch pair, the SA-VAR model is used to make the self-prediction and the cross-prediction for the reference patch, which results in two complexity terms, i.e., the self-complexity of the reference patch and the transformational complexity between the local patch pair. The prediction terms generated by the SA-VAR serve as complementary terms to help generate local quality indices. Finally, we fuse all local quality indices to get the final quality score.  

\begin{figure}
    \centering
    \subfigure[]{\includegraphics[width=\linewidth]{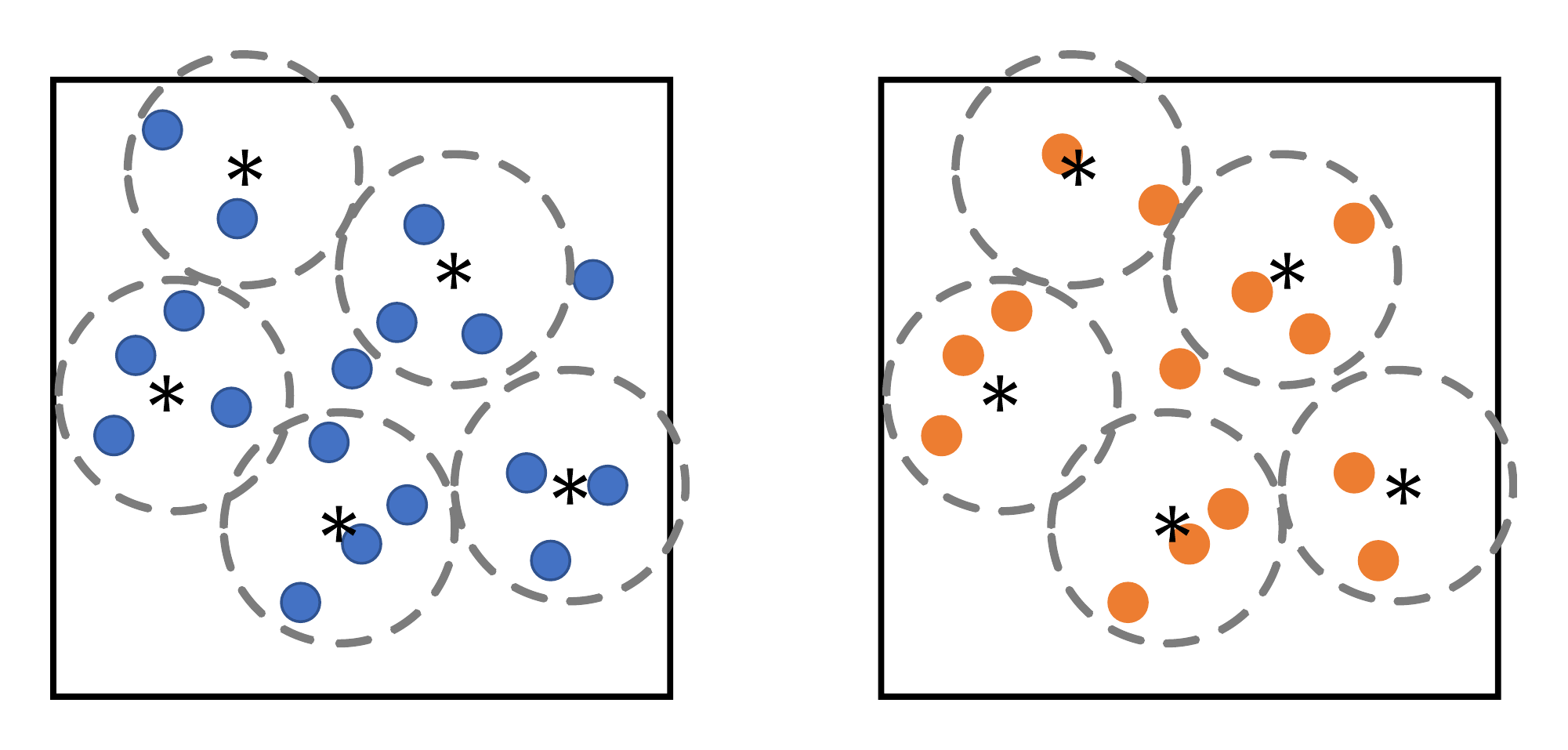}}
    \subfigure[]{\includegraphics[width=\linewidth]{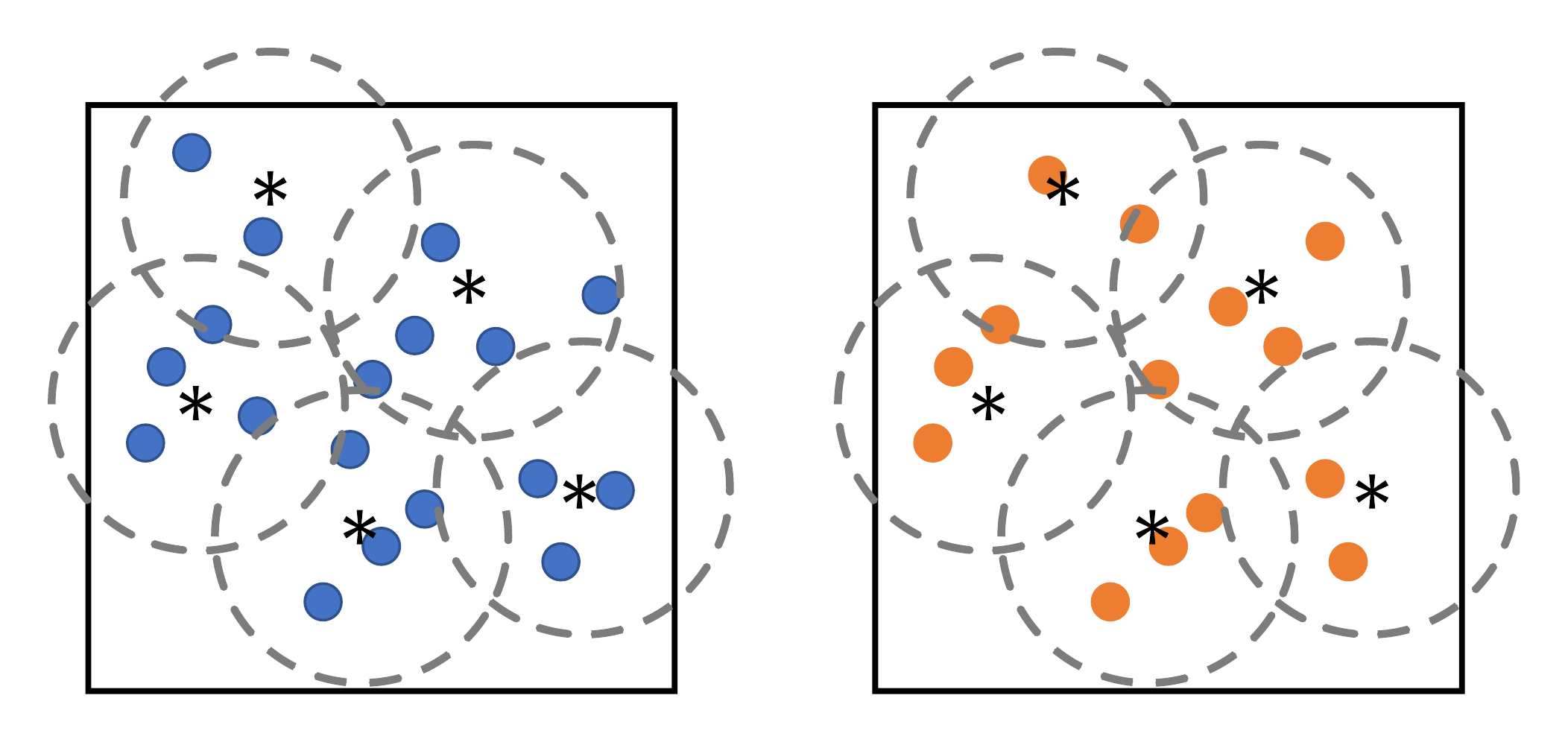}}
    \subfigure[]{\includegraphics[width=\linewidth]{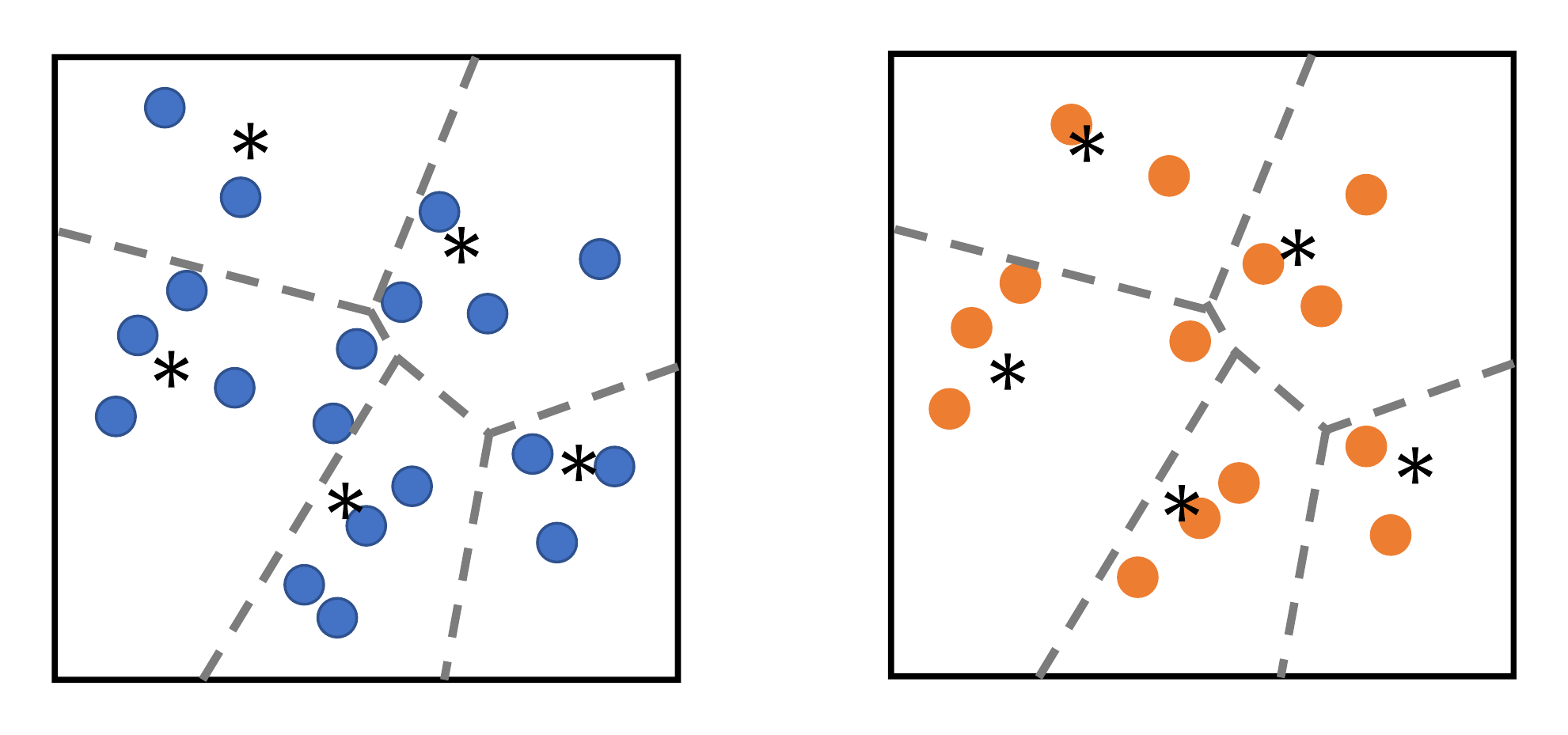}}

    \caption{The comparison of several segmentation strategies for the FR-PCQA task. Note that the "*" denotes the spherical center in (a) and (b) and the generating seeds in (c). (a) Sphere-based segmentation with a small radius; (b) sphere-based segmentation with a large radius; 
    (c) space segmentation using the Voronoi diagram.}
    \label{fig:cluster_comparison}
\end{figure}

\subsection{Space Segmentation} \label{sec:segmentation}
It is preferred to process the point cloud locally rather than globally in PCQA. First, dense point clouds have millions of disordered points distributed in 3D space, which leads to high computational cost when taking the entire point cloud as input. Especially, when the complexity of one algorithm reaches $O(N^2)$ or higher, the practical computational cost of global processing will be prohibitive. Second, due to the multi-view characteristic when observing 3D objects and the foveation mechanism \cite{lee2002foveated} of the HVS, human observers often only perceive a localized area of the point cloud at one time. Finally, \znote{local processing can help differentiate the intrinsic characteristics of different regions to better exploit some visual properties such as the masking effect \cite{hua2022cpc}. } 

For the FR-PCQA task, two critical steps need to be determined for the local processing, i.e., how to divide the point cloud into local patches and how to construct the patch correspondence between the reference and distorted point clouds. Existing methods usually try to bound points in the reference and its distorted version by a sequence of spherical spaces \cite{yang2020inferring},\cite{zhang2021ms}. However, the number of divisions and the size of the spherical radius in the method are often difficult to determine. As illustrated in Fig. \ref{fig:cluster_comparison} (a) and (b), given a fixed number of divisions, a small radius is not enough to cover the entire point cloud, while a large radius results in many overlaps and involves unnecessary computational cost. Some researchers \cite{hua2022cpc} leverage $K$-means clustering for the reference and then determine the corresponding clusters of the distorted version by the nearest-neighbor searching, which may cause some points in the distorted point cloud to be ignored. To avoid the above problems, our work uses a 3D Voronoi diagram to realize the space segmentation for point clouds, which can achieve non-overlapping divisions and avoid the omission of points during the process of correspondence construction.    

The VD divides the 3D space into many non-overlapping convex polyhedrons called Voronoi cells (VCs) so that each polyhedron contains exactly one generating seed, and every point in a given cell is much closer to the corresponding generating seed than to any other generating seeds. Let $\p\in\R^3 $ be a point in the Euclidean space. Assuming a set of points as the generating seeds $\S=\{\s_l \in \R^3\}_{l=1}^{L}$, the VD generated by $\S$ is defined by the nearest–neighbor principle, i.e., the cell {$VC_l$} corresponding to $\s_l$ is defined by 
\begin{equation}\label{eq:VC}
    VC_l = \{\p\in\R^3|\rm{dist}(\p,\s_l)\leqslant \rm{dist}(\p,\s_k), if\quad k\neq l\},
\end{equation}
where $\rm{dist}(\cdot)$ denotes the Euclidean distance between two points. Clearly, one cell $VC_l$ represents a convex polyhedron space centered around the seeds $\s_l$ bounded by multiple planes, which is locally finite and has a non–empty interior. All VCs cover the whole 3D space without overlap and omission. Hence, we can bound points in both the reference and distorted point clouds via these cells.

Specifically, let $\X = \{\x_i\in\R^6\}_{i=1}^N$ and $\Y = \{\y_j\in\R^6\}_{j=1}^M$ be the reference and distorted point clouds with $N$ and $M$ points, respectively. We first derive the generating seed set $\S$ from $\X$ via the farthest point sampling (FPS), i.e., 
\begin{equation}
    \S=\lfloor\Upsilon(\X)\rfloor_{L} \in \R^{L\times6}, L\ll N,\nonumber
\end{equation}
where $\Upsilon(\cdot)$ represents the FPS operation. The number of seeds is empirically set as $L=400$ in our work.

Next, for each seed $\s_l$, we derive its corresponding cell $VC_l$ according to Eq. \eqref{eq:VC}. Then we can respectively obtain the local patch pair bounded in $\X$ and $\Y$ by the $VC_l$ as
\begin{equation}
\begin{aligned}
    \X_l&=\{\x_i|\x_i^O\in VC_l\}\in \R^{N_l\times6},\\
    \Y_l&=\{\y_j|\y_j^O\in VC_l\}\in \R^{M_l\times6},
\end{aligned}
\end{equation}
where $N_l$ and $M_l$ denote the point number of two local patches. In total, we have $L$ local patch paires derived from $\X$ and $\Y$. Note that all patches in $\X$ or $\Y$ are non-overlapping and no point is omitted during the above process.  

Finally, considering relative coordinates are more meaningful than absolute coordinates when human eyes focus on one specific region, we align patches via translation so that each patch has its generating seed at the origin, i.e.,
\begin{equation}
\begin{aligned}
    \Xt_l &=\{\xt_i\}_{i=1}^{N_l}= [\X_l^O - \S_{N_l},\X_l^I]\in \R^{N_l\times6},\\
    \Yt_l &=\{\yt_j\}_{j=1}^{M_l}=[\Y_l^O - \S_{M_l},\Y_l^I]\in \R^{M_l\times6},
\end{aligned}
\end{equation}
where $\xt_i = [\xt_i^O,\xt_i^I]=[\x_i^O-\s_l^O,\x_i^I]$ denotes the translated point of $\x_i$ and $\S_{N_l}=[(\s_l^O)_{\times N_l}]\in\R^{N_l\times3}$ denotes the repeated coordinate matrix of $\s_l$. $\yt_j$ and $\S_{M_l}$ have a similar meaning as above. The translated local patch pair, i.e., $\Xt_l$ and $\Yt_l$, is used for subsequent local processing.

\subsection{Complexity-based Feature} \label{sec:SA-VAR}
As mentioned in Section \ref{sec:trans_complexity_theory}, the predictive coding theory shares a similar principle of redundancy reduction with information compression and has a good connection to visual perception, which makes it an effective theorem basis for measuring the complexity of data. In general, our basic idea is to relate the complexity to the negative log-likelihood function after minimization.

An obvious fact is that any quantitative application of Eq. \eqref{eq:optimal_theta} needs to assume the brain generative model $G$ first. A model with higher expressive power might better approximate the brain, but incurs higher computational complexity. In some previous studies \cite{zhai2011psychovisual,wu2012perceptual, zhang2008image, wu2010adaptive, gu2017evaluating}, the autoregressive model (AR) has been widely used due to its good local adaptability and low computational cost. The classic paradigm of the AR model used in the image domain can be written as 
\begin{equation}\label{eq:image_ar}
    f_i = \sum_{j=1}^K\theta_j f_j+e_i=f(\mathcal{N}^K_{i})\thetaset+e_i,
\end{equation}
where $f_i$ represent the $i$-th pixel value in the given image; $f(\mathcal{N}^k_{i})\in\R^K$ is a vector consisting $K$ nearest neighbors of the $i$-th pixel and $e_i$ is the additive Gaussian residual noise with zero mean. The above AR model achieves satisfying performance in many image processing algorithms (e.g., interpolation \cite{zhang2008image}, compression \cite{wu2010adaptive}, quality assessment \cite{zhai2011psychovisual}, etc.).

However, it is worth pointing out that the AR model in Eq. \eqref{eq:image_ar} only concentrates on a single channel (e.g., luminance channel in the image domain), which is effective enough for regular single-channel data, such as gray images. In comparison, the colored point cloud studied in this paper is typical multi-channel data and there exists some correlation among different feature channels (e.g., XYZ and RGB channels) \cite{irfan2021joint}. It is believed that our HVS can capture such a correlation to realize the final quality decision. Therefore, in order to exploit the inter-channel and intra-channel correlation of point cloud data simultaneously, we propose an SA-VAR model to encode multiple channels.

Specifically, given one translated local patch pair $\Xt_l$ and $\Yt_l$,
we calculate the self-complexity $C(\Xt_l|\Xt_l)$ and the transformational complexity $C(\Xt_l|\Yt_l)$. Here, we denote the self-complexity by $C(\Xt_l|\Xt_l)$ rather than $C(\Xt_l)$ because the AR model can be regarded as a self-generation model when applied only to $\Xt_l$ itself. We detail the calculation process of the above two terms in the following.

\begin{figure}
    \centering
    \subfigure[]{\includegraphics[width=0.42\linewidth]{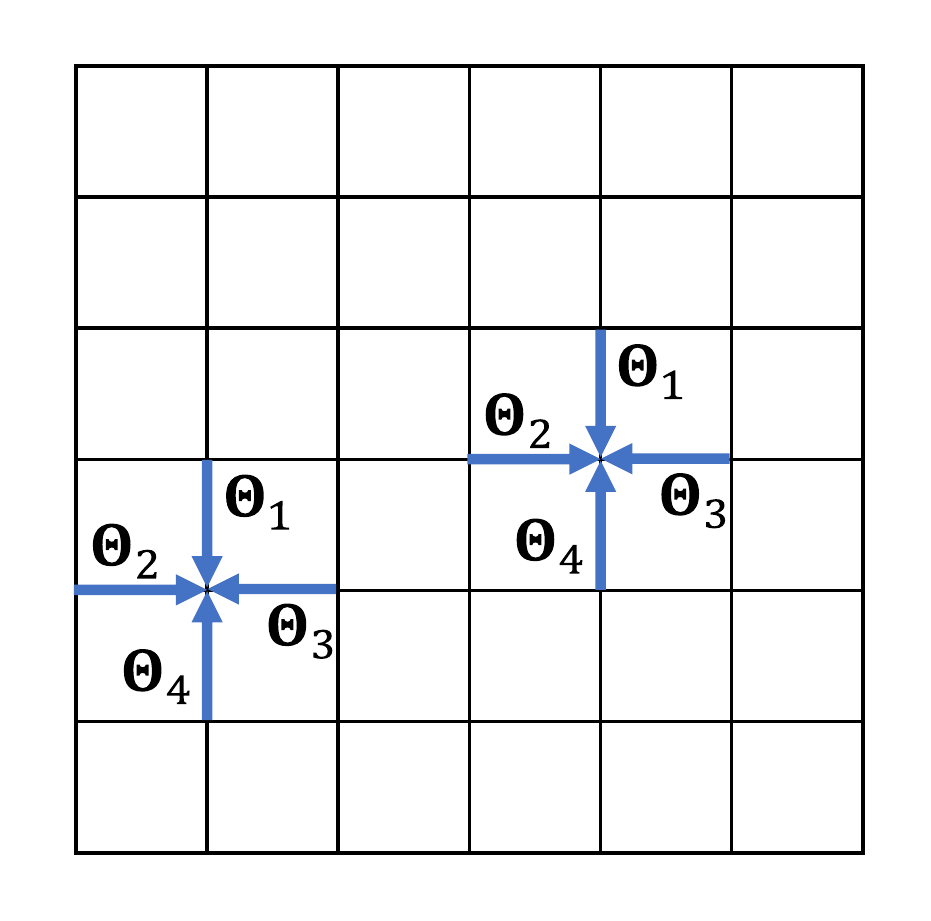}}
    \subfigure[]{\includegraphics[width=0.55\linewidth]{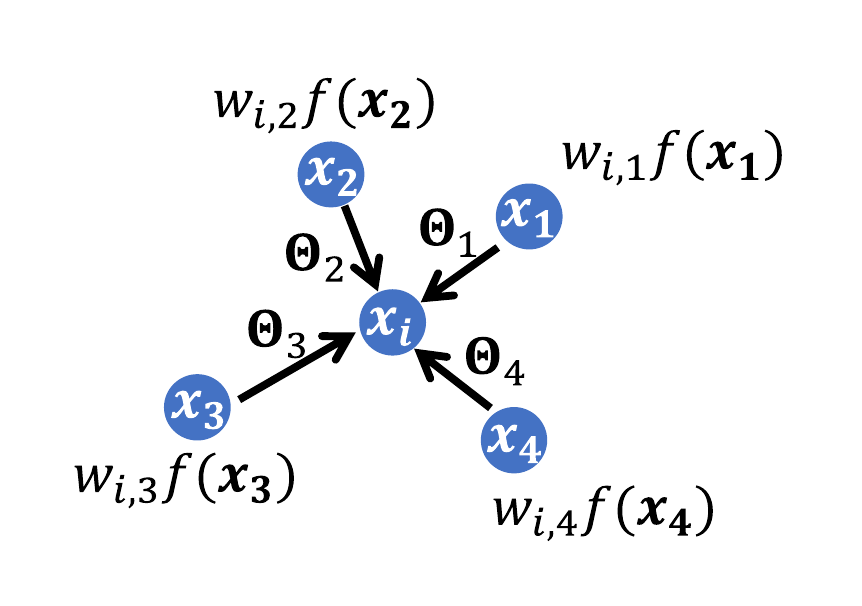}}

    \caption{Illustration of different AR models. (a) 2D AR model; (b) SA-VAR model.}
    \label{fig:spatial weight}
\end{figure}

\textbf{Calculation of the Self-Complexity.}
We first show the classic VAR model for encoding multiple variables in $\Xt_l$. For one point $\xt_i$ in $\Xt_l$, the VAR leverages the neighbors of $\xt_i$ in $\Xt_l$ to make the self-prediction for its feature, i.e.,
\begin{equation}\label{eq:VAR}
    f(\xt_i) = \sum\nolimits_{\xt_j\in\NxX}\Thetaset_{j}f(\xt_j)+\e_i=\Thetaset_s f(\NxX)+\e_i,
\end{equation}
where $f(\xt_i)\in\R^d$ represents one $d$-dimension feature of $\xt_i$, accounting for $\xt_i^O$ or $\xt_i^I$ in our work. In other words, we utilize the VAR to encode the XYZ and RGB channels individually.  $\NxX \subset\Xt_l$ denotes $\xt_i$'s $K$ closest points in $\Xt_l$ (note that the closest point, i.e., $\xt_i$ itself, is excluded) and $f(\NxX)=[f(\xt_1);f(\xt_2);\cdots;f(\xt_K)]\in\R^{Kd}$ denotes the feature matrix of the neighbors. $\Thetaset_j\in\R^{d\times d}$ and $\Thetaset_s=[\Thetaset_1;\Thetaset_2;\cdots;\Thetaset_K]\in\R^{d\times Kd}$ denote the parameter set of VAR. $\e_j\sim \mathcal{N}(0,\Sigma_{s})$ is assumed to belong to multi-variate Gaussian distribution, where $\Sigma_{s}\in\R^{d\times d}$.

As shown in Fig. \ref{fig:spatial weight} (a), the existing assumption of traditional 2D AR is that all points/pixels are located in a regular grid shape \cite{takeuchi2017autoregressive}, in which case the parameter $\Thetaset_s$ shares the same spatial structure in different locations. In comparison, the topological condition of the point cloud is irregular as shown in Fig. \ref{fig:spatial weight} (b). To adapt the VAR to point cloud data, we add a spatial weight for each neighbor in Eq. \eqref{eq:VAR} to produce one  SA-VAR model, i.e., 
\begin{equation}\label{eq:SA-VAR}
\begin{aligned}
\begin{split}
    f(\xt_i) &= \sum\nolimits_{\xt_j\in\NxX}w_{i,j}\Thetaset_{j}f(\xt_j)+\e_i \\
    &=\Thetaset_s f_w(\NxX)+\e_i, \\
    w_{i,j}&=\frac{d_{i,j}}{\sum\nolimits_j{d_{i,j}}},d_{i,j}=\frac{1}{1+e^{-\frac{\|\xt_i^O-\xt_j^O\|_2}{\eta}}},
\end{split}
\end{aligned}
\end{equation}
where $\eta$ denotes the geometry variance within $\NxX$. The weight $w_{i,j}$ is related to the spatial distance between $\xt_i$ and $\xt_j$, which strengthens the AR's ability to capture irregular spatial structure. 

Now we apply the above SA-VAR for all points in the local patch $\Xt_l=\{\xt_i\}_{i=1}^{N_l}$ and have
\begin{equation}\label{eq:SA-VAR matrix}
    f(\Xt_l)=\Thetaset_s f_w(\NXX)+\E,
\end{equation}
where $f(\Xt_l)$, $\E \in\R^{d\times N_l}$ denote the feature and noise matrix of $\Xt_l$; $f_w(\NXX)\in\R^{Kd\times N_l}$ denotes the weighted neighbor feature matrix of the whole patch $\Xt_l$. To facilitate the analysis, we vectorize the above representation as 
\begin{equation}\label{eq:SA-VAR vec}
    f_v(\Xt_l)=(f_w^T(\NXX)\otimes\I_d)\Thetaset_{s,v}+\E_v,
\end{equation}
where $f_v(\Xt_l)$, $\E_v\in\R^{N_ld}$ and $\Thetaset_{s,v}\in\R^{Kd^2}$ denote the vectorized representation of $f(\Xt_l)$, $\E$ and $\Thetaset_s$; $\otimes$ represents the Kronecker product; $\I_d\in\R^{d\times d}$ is the $d$-dimension identity matrix. Considering $\E_v\sim\mathcal{N}(0,\I_{N_l}\otimes\Sigma_s)$, we have the likelihood 
\begin{equation}
\begin{aligned}
    p(\Xt_l|\Thetaset_{s,v},G)&=\frac{1}{(2\pi)^{\frac{N_ld}{2}}|\I_{N_l}\otimes\Sigma_s|^\frac{1}{2}}e^{-\frac{1}{2}\E_v^T(\I_{N_l}\otimes\Sigma_s)^{-1}\E_v}.
\end{aligned}
\end{equation}
The negative log-likelihood term can be further written as 
\begin{equation}\label{eq:loglikelihood}
\begin{aligned}
    -\log{p(\Xt_l|\Thetaset_{s,v},G)}&=\frac{N_ld}{2}\log{2\pi}+\frac{N_l}{2}\log{|\Sigma_s|}\\
    &+\frac{1}{2}\Tr(\E^T\Sigma_s^{-1}\E).
\end{aligned}
\end{equation}
According to Eq. \eqref{eq:optimal_theta} and Eq. \eqref{eq:loglikelihood},  we can derive the closed-form solution of $\Thetaset_{s,v}$ and $\Sigma_s$ as 
\begin{small}
\begin{equation}
\begin{aligned}
    \hat{\Thetaset}_{s,v}&=[(f_w(\NXX)f_w^T(\NXX))^{-1}f_w(\NXX)\otimes\I_d]f_v(\Xt_l),\\
    \hat{\Sigma}_s &=\frac{1}{N_l}[f(\Xt_l)-\hat{\Thetaset}_sf_w(\NXX)][f(\Xt_l)-\hat{\Thetaset}_s f_w(\NXX)]^T,
\end{aligned}\label{eq:minimized_solution}
\end{equation}
\end{small}
where $\hat{\Thetaset}_s$ denotes the reshaped matrix form of $\hat{\Thetaset}_{s,v}$.

Inducing the optimal estimation $\hat{\Thetaset}_{s,v}$ and $\hat{\Sigma}_s$ in Eq. \eqref{eq:loglikelihood}, we obtain the minimized negative log-likelihood function as:
\begin{equation}\label{eq:diff_entropy}
    -\log{p(\Xt_l|\hat{\Thetaset}_{s,v},G)}=\frac{N_ld}{2}(\log{2\pi}+1)+\frac{N_l}{2}\log{|\hat{\Sigma}_s|}.
\end{equation}
Eq. \eqref{eq:diff_entropy} represents the differential entropy of the multi-variate Gaussian distribution, which is only relevant to the determinant of the estimated covariance matrix when fixing $N_l$ and $d$. Therefore, we represent the self-complexity of $\Xt_l$ as
\begin{equation}\label{eq:self complexity}
    C(\Xt_l|\Xt_l)=|\hat{\Sigma}_s|.
\end{equation}
In practice, we apply the SA-VAR for the XYZ and RGB channels, respectively, in which cases we have $f(\Xt_l)=(\Xt_l^O)^T$ and $f(\Xt_l)=(\Xt_l^I)^T$. Finally, we represent the self-complexity of geometry and color channels using $C^O(\Xt_l|\Xt_l)$ and $C^I(\Xt_l|\Xt_l)$.

\textbf{Calculation of  the Transformational Complexity.}\label{sec:cal of tc}
We further consider the transformational complexity between $\Xt_l$ and $\Yt_l$. As mentioned in Section \ref{sec:trans_complexity_theory}, the transformational complexity $C(\Xt_l|\Yt_l)$ can be considered as the code length of $\Xt_l$ when $\Yt_l$ is given. Therefore, we utilize the neighbors of $\xt_i$ in $\Yt_l$ to make the cross-prediction for $\xt_i$, i.e.,
\begin{equation}
\begin{aligned}
    f(\Xt_l)&=\Thetaset_t f_w(\NXY)+\E,\\
    f_v(\Xt_l)&=(f_w^T(\NXY)\otimes\I_d)\Thetaset_{t,v}+\E_v,\\
\end{aligned}
\end{equation}
where $\NXY \subset\Yt_l$ has a similar meaning to $\NXX$. $\E_v\sim\N(0,\I_{N_l}\otimes\Sigma_t)$. The likelihood function can be written as 
\begin{equation}\label{eq:X2Y likelihood}
    p(\Xt_l|\Yt_l,\Thetaset_{t,v},G)=\frac{1}{(2\pi)^{\frac{N_l d}{2}}|\I_{N_l}\otimes\Sigma_t|^\frac{1}{2}}e^{-\frac{1}{2}\E_v^T(\I_{N_l}\otimes\Sigma_t)^{-1}\E_v}.
\end{equation}  

According to Eq. \eqref{eq:loglikelihood}-\eqref{eq:self complexity}, the minimized negative log-likelihood term of Eq. \eqref{eq:X2Y likelihood} is only related to the optimized covariance matrix. Therefore, we represent the transformational complexity between $\Xt_l$ and $\Yt_l$ as
\begin{equation}\label{eq:X2Y complexity}
    C(\Xt_l|\Yt_l) = |\hat{\Sigma}_t|,
\end{equation}
where $\hat{\Sigma}_t$ can be obtained via the same paradigm as Eq. \eqref{eq:minimized_solution}.  Similarly, we respectively derive $C^O(\Xt_l|\Yt_l)$ and $C^I(\Xt_l|\Yt_l)$, which denotes the transformational complexity of geometry and color channels. 

\textbf{Pooling Strategy.}
As illustrated in Section \ref{sec:trans_complexity_theory}, we need to choose a pooling strategy to merge the self-complexity and the transformational complexity. Considering the success of the similarity measurement in SSIM \cite{wang2004image}, we define two local complexity-based features related to geometry and color channels as

\begin{equation}
\begin{aligned}
    F^O_{1,l} =&\frac{2C^O(\Xt_l|\Xt_l)\cdot C^O(\Xt_l|\Yt_l)+T}{(C^O(\Xt_l|\Xt_l))^2+(C^O(\Xt_l|\Yt_l))^2+T}, \\  
    F^I_{1,l} =&\frac{2C^I(\Xt_l|\Xt_l)\cdot C^I(\Xt_l|\Yt_l)+T}{(C^I(\Xt_l|\Xt_l))^2+(C^I(\Xt_l|\Yt_l))^2+T},
\end{aligned}
\end{equation}
where $T$ is a small no-zero constant to prevent numerical instability.

\begin{figure*}
    \centering
    \includegraphics[width=0.85\linewidth]{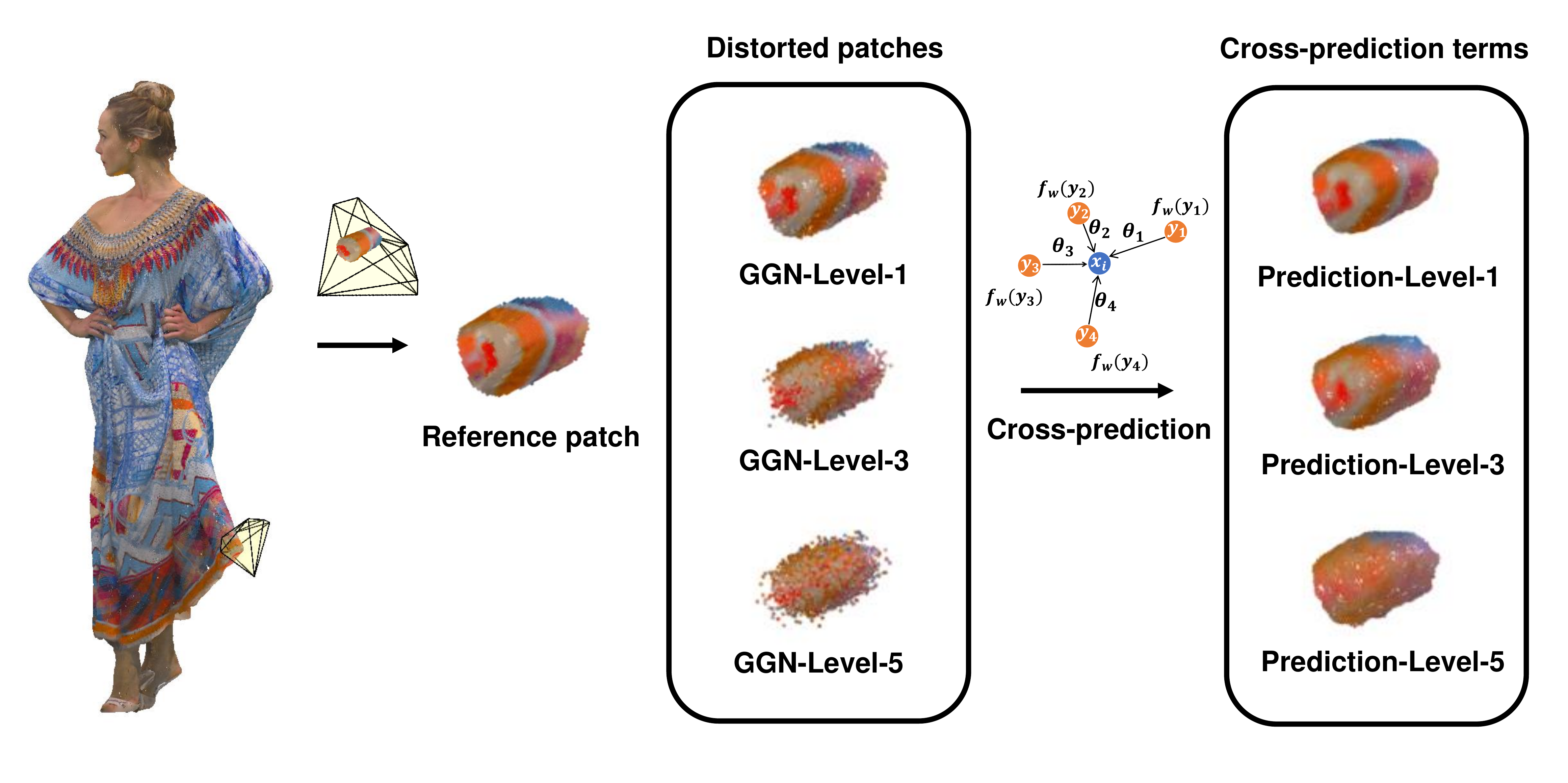}
    
    \caption{Comparison of the cross-prediction terms of distorted patches impaired by GGN at different levels. As the level of distortion increases, the reconstructed results (referred to as cross-prediction terms) of the distorted patches exhibit greater disparity from the reference patch.}
    \label{fig:prediction_term}
\end{figure*}

\subsection{Prediction-based Feature}
Except for complexity-based terms shown in Eq. \eqref{eq:self complexity} and \eqref{eq:X2Y complexity}, the prediction terms generated by SA-VAR are also meaningful. Specifically, we define
\begin{equation}
\begin{aligned}
    \Xhat_l &= \{\xhat_i\}_{i=1}^{N_l}= [\hat{\Thetaset}_s f_w(\NXX)]^T\in\R^{N_l\times 6},\\
    \Yhat_l &= \{\yhat_i\}_{i=1}^{N_l}= [\hat{\Thetaset}_t f_w(\NXY)]^T\in\R^{N_l\times 6},
\end{aligned}
\end{equation}
which represent the reconstructed patches from applying the SA-VAR for the self-prediction of $\Xt_l$ and for the cross-prediction between $\Xt_l$ and $\Yt_l$. 
We show the cross-prediction terms of GGN at different levels in Fig. \ref{fig:prediction_term}.
Two reasons motivate us to utilize the above prediction terms: i) there exists a perfect \textit{point-to-point correspondence} between $\Xhat_l$ and $\Yhat_l$ because both $\xhat_i$ and $\yhat_i$ denote the predictions for $\xt_i$, which is highly desirable for point cloud processing; ii) according to the predictive coding theory, the prediction terms can be regarded as the optimized perceptual representation in our brain \cite{wu2012perceptual}, which also reflects the visual quality to some extent. For instance, in Fig. \ref{fig:prediction_term}, we can observe that the cross-prediction term of the patch at the highest GGN level does not reproduce the patch shape and color well, which can be explained by the fact that too disorderly points bring high difficulty for our brain to achieve perfect reconstruction. Moreover, the formulation of the complexity terms is related to the residual errors between the prediction and the ground truth. Therefore, it is believed that the prediction terms are complementary to the complexity terms for modeling the overall perception of point cloud distortions.

Let the point index of $\Xhat_l$ and $\Yhat_l$ be from 1 to $N_l$. To better exploit the point-to-point correspondence between $\Xhat_l$ and $\Yhat_l$, we first calculate one point-wise difference vector between $\xhat_i$ and its neighbors in $\Xhat_l$  as follows,
\begin{equation}\label{eq:F2_step1}
\begin{aligned}
    \v_{\xhat_i} &= \{g(\xhat_i, \xhat_j)\}_{j\in {Id}_i}\in\R^K,\\
    g(\xhat_i, \xhat_j) &= (\sum_{d}^3  k_d |(\xhat_i^I)_j - (\xhat_j^I)_j|+1)\cdot \|\xhat_i^O-\xhat_j^O\|_2,
\end{aligned}
\end{equation}
where ${Id}_i=\{j|\xhat_j\in\NxXhat\}$ denotes the indices of $\xhat_i$'s neighbors; $k_d$ represents the weighting factors between different color channel,  i.e., $k_R: k_G: k_B = 1:2:1$ \cite{yang2021mped}. $g(\cdot)$ can be seen as a combination of the geometry difference used in p2po and the color difference used in $\rm{PSNR}_{YUV}$, reflecting both the geometric and attribute discrepancies between one point pair.

Then we utilize $Id_i$ to find the corresponding point in $\Yhat_l$ and compute the point-wise difference vector between $\yhat_i$ and these points as 
\begin{equation}
    \v_{\yhat_i} = \{g(\yhat_i,\yhat_j)\}_{j\in Id_i}\in\R^K.
\end{equation}
Repeating the above step for all points in $\Xhat_l$ and $\Yhat_l$, we have $V_{\Xhat_l}=[\v_{\xhat_1};\v_{\xhat_2};\cdots;\v_{\xhat_{N_l}}]\in\R^{N_l K}$ and
$V_{\Yhat_l}=[\v_{\yhat_1};\v_{\yhat_2};\cdots;\v_{\yhat_{N_l}}]\in\R^{N_l K}$. The two vectors reflect the local variation of two patches. Similarly to some previous studies\cite{yang2020inferring}, \cite{wang2004image}, we derive the variance and of $\V_{\Xhat_l}$ and $\V_{\Yhat_l}$ ( denoted by $\sigma_{{\Xhat_l}}$ and $\sigma_{{\Yhat_l}}$) and the covariance between two vectors (denoted by $c_{\Xhat_l,\Yhat_l}$). Finally, we compute an auxiliary feature related to the prediction term as 
\begin{equation}
    F_{2,l} =\frac{c_{\Xhat_l,\Yhat_l}+T}{\sigma_{{\Xhat_l}}\cdot\sigma_{{\Yhat_l}}+T}.
\end{equation}

\subsection{Calculation of Visual  Quality Score} \label{sec:score_pooling}
For each local patch pair $\X_l$ and $\Y_l$, we have two complexity-based features $F^O_{1,l}$ and $F^I_{1,l}$ and one prediction-based feature $F_{2,l}$, which are both obtained via the SA-VAR. We then fuse these local indices together to have two global indices for the entire distorted point cloud as 
\begin{equation}\label{eq:feature_fusion}
\begin{aligned}
    F_1 &= \underbrace{(\frac{1}{L}\sum\nolimits_l{F^O_{1,l}})}_{F_1^O}\cdot\underbrace{(\frac{1}{L}\sum\nolimits_l{F^I_{1,l}})}_{F_1^I}, \\
    F_2 &= \frac{1}{L}\sum\nolimits_l{F_{2,l}}.
\end{aligned}
\end{equation}

In the end, we can have the overall point cloud quality score by weighting the above global indices as
\begin{equation}
    Q=\alpha F_1+(1-\alpha) F_2,
\end{equation}
where $\alpha\in[0,1]$ is the weighting factor.

\begin{table*}[]
\caption{PERFORMANCE COMPARISON OF DIFFERENT FR-PCQA METRICS ON FIVE DATABASES}
\label{tab:overall_performance}
\resizebox{181mm}{!}{
\setlength{\tabcolsep}{2.5pt}
\renewcommand\arraystretch{1.2}
\begin{tabular}{l|ccc|ccc|ccc|ccc|ccc|ccc}
\hline
Database & \multicolumn{3}{c|}{SJTU-PCQA\cite{yang2020predicting}} & \multicolumn{3}{c|}{WPC\cite{liu2022perceptual}} & \multicolumn{3}{c|}{M-PCCD\cite{alexiou2019comprehensive}} & \multicolumn{3}{c|}{ICIP2020\cite{perry2020quality}} & \multicolumn{3}{c|}{IRPC\cite{javaheri2020point}} &\multicolumn{2}{c}{Average ranking}  \\ \hline
Criteria & PLCC & SROCC & RMSE & PLCC & SROCC & RMSE & PLCC & SROCC & RMSE & PLCC & SROCC & RMSE & PLCC & SROCC & RMSE  &PLCC & SROCC \\\hline
MSE-p2po \cite{MPEGSoft} &0.877	&0.791	&1.166	&0.578	&0.566	&18.708	&0.778	&0.797	&0.855	&0.888	&0.878	&0.522	&0.769	&0.714	&0.625	&6.4	&6.4
\\
MSE-p2pl \cite{MPEGSoft} &0.753	&0.676	&1.596	&0.488	&0.446	&20.013	&0.815	&0.836	&0.788	&0.913	&0.915	&0.463	&0.749	&0.682	&0.647	&6.8	&7.2
\\
Hausdorff-p2po \cite{MPEGSoft} &0.742	&0.681	&1.628	&0.398	&0.258	&21.028	&0.593	&0.366	&1.095	&0.601	&0.542	&0.908	&0.401	&0.044	&0.895	&9.8	&10.4
\\
Hausdorff-p2pl\cite{MPEGSoft} &0.737	&0.670	&1.639	&0.383	&0.315	&21.171	&0.571	&0.507	&1.117	&0.649	&0.602	&0.865	&0.144	&0.252	&0.967	&10.4	&10.0
\\
$\rm PSNR_{YUV}$ \cite{MPEGSoft}&0.652	&0.644	&1.841	&0.551	&0.536	&19.132	&0.654	&0.660	&1.029	&0.868	&0.867	&0.564	&0.721	&0.673	&0.677	&8.8	&8.6
\\
PCQM \cite{meynet2020pcqm} &0.860	&0.847	&1.237	&\bf{0.751}	&\bf{0.743}	&\bf{15.132}	&0.900	&0.915	&0.594	&\bf{0.969}	&\bf{0.970}	&\bf{0.280}	&0.897	&0.831	&0.433	&\bf{3.2}	&\bf{3.0} 
\\
GraphSIM \cite{yang2020inferring} &0.856	&0.841	&1.071	&0.694	&0.680	&16.498	&\bf{0.932}	&\bf{0.945}	&\bf{0.493}	&0.890	&0.872	&0.518	&0.850	&0.716	&0.514	&5.0	&4.4
\\
MS-GraphSIM \cite{zhang2021ms} &\bf{0.897}	&{0.874}	&\bf{1.071}	&0.717	&0.707	&15.970	&0.916	&0.930	&0.545	&0.906	&0.895	&0.481	&0.879	&0.720	&0.466	&3.6	&3.6
\\
pointSSIM \cite{alexiou2020towards} &0.725	&0.704	&1.672	&0.510	&0.454	&19.713	&0.926	&0.918	&0.514	&0.904	&0.865	&0.486	&0.636	&0.578	&0.754	&7.2	&7.4
\\
MPED \cite{yang2021mped} &0.896	&\bf{0.884}	&1.076	&0.700	&0.678	&16.374	&0.824	&0.849	&0.770	&\bf{0.964}	&\bf{0.952}	&\bf{0.303}	&\bf{0.923}	&\bf{0.842}	&\bf{0.376}	&\bf{3.2}	&3.2
\\ \hline

TCDM &\bf{0.930}	&\bf{0.910}	&\bf{0.891}	&\bf{0.807}	&\bf{0.804} &\bf{13.525}	&\bf{0.936}	&\bf{0.944}	&\bf{0.479}	&0.942	&0.935	&0.382	&\bf{0.921}	&\bf{0.840}	&\bf{0.380}	&\bf{1.6}	&\bf{1.8}	\\ \hline
	
\end{tabular}}
\end{table*}

\begin{figure*}
    \centering
    \subfigure[]{\includegraphics[width=0.3\linewidth]{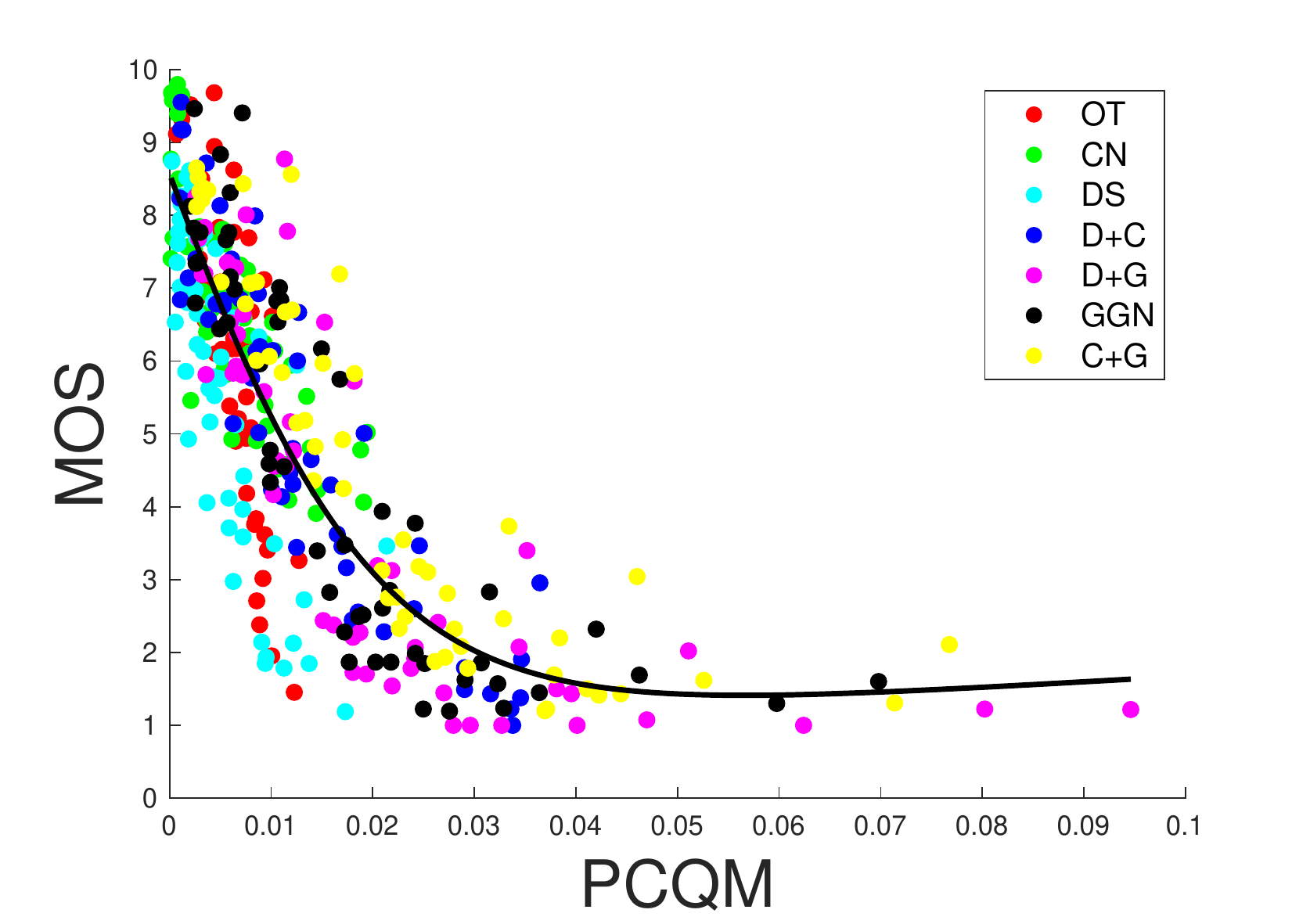}}
    \subfigure[]{\includegraphics[width=0.3\linewidth]{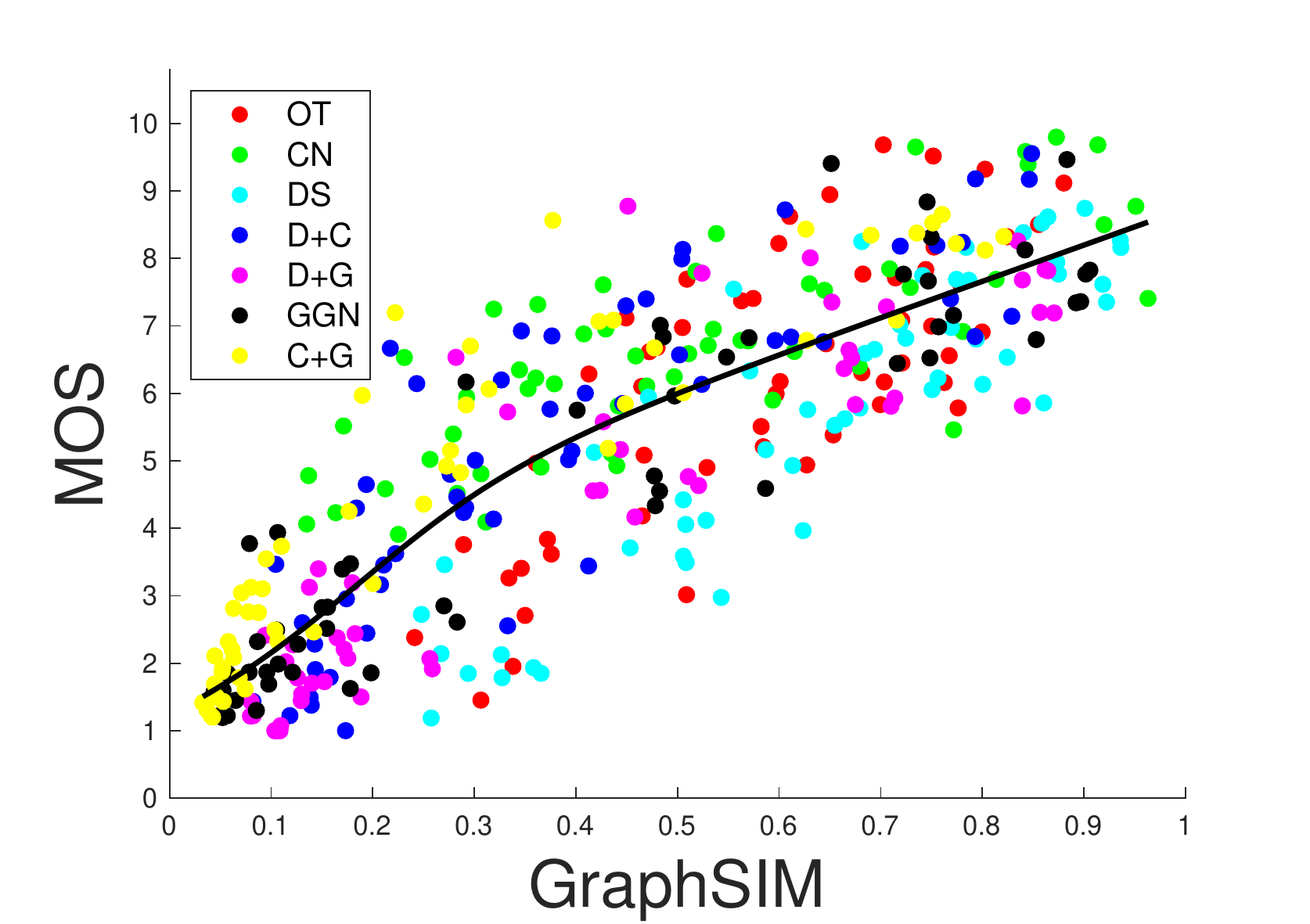}}
    \subfigure[]{\includegraphics[width=0.3\linewidth]{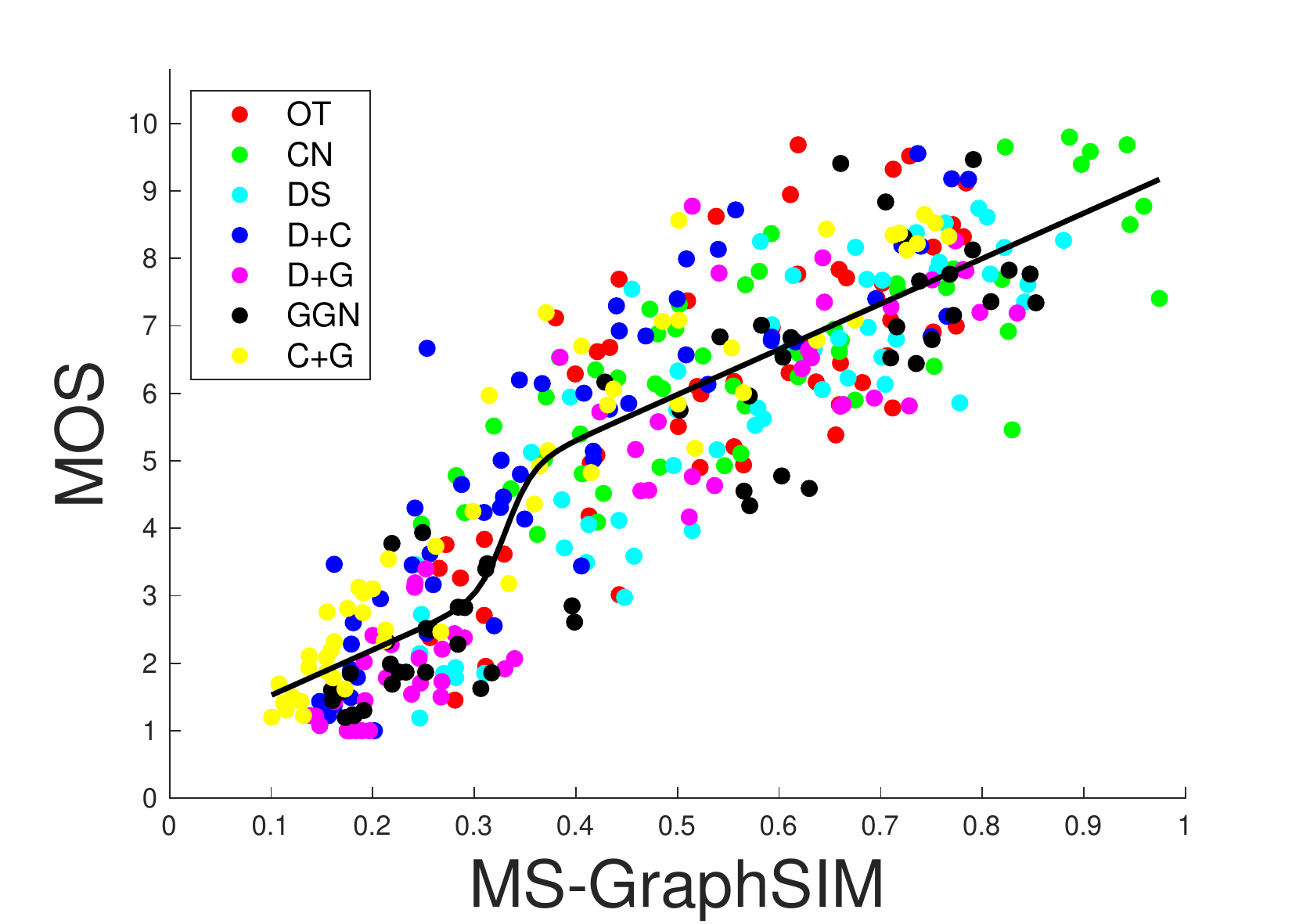}}
    \subfigure[]{\includegraphics[width=0.3\linewidth]{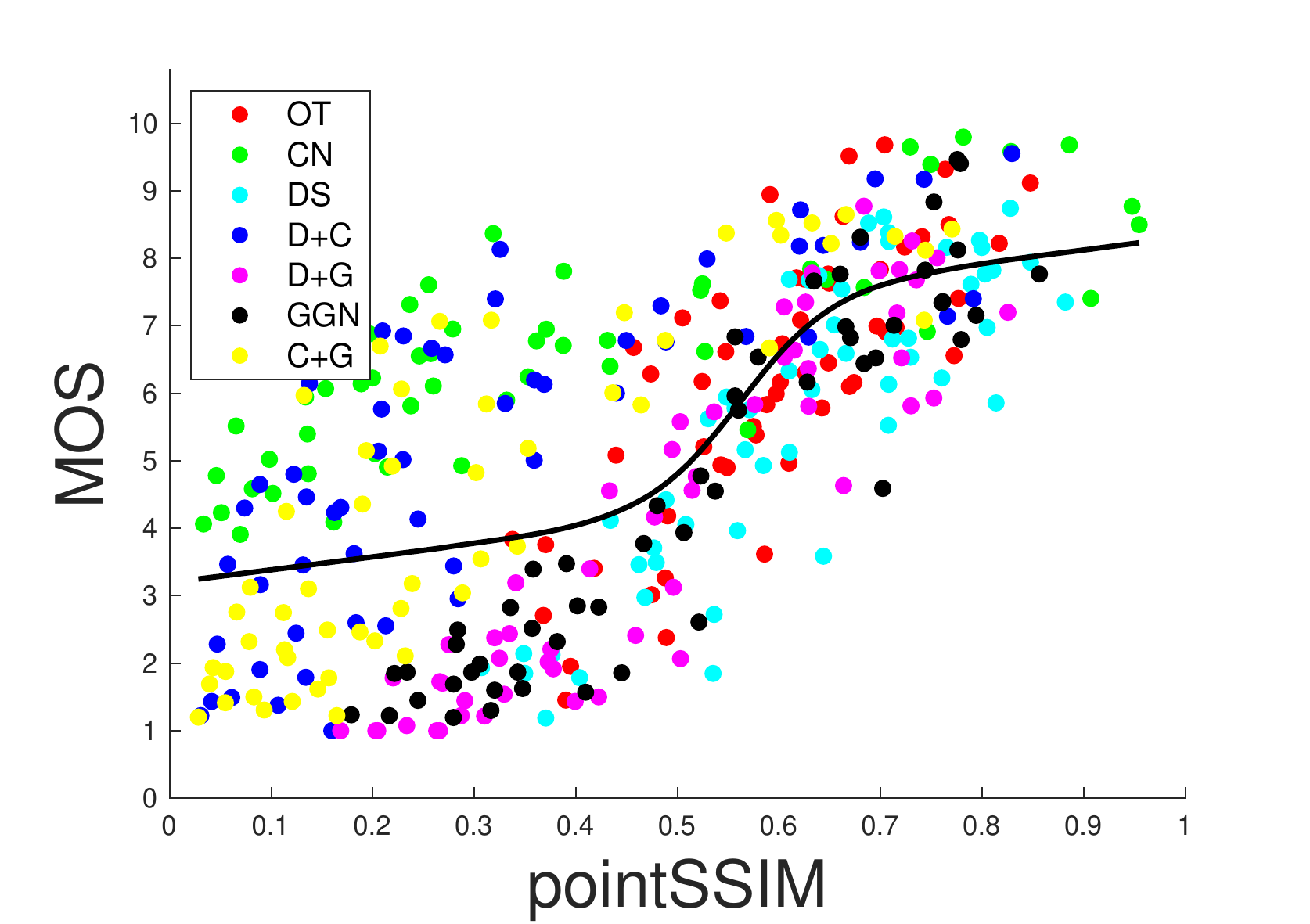}}
    \subfigure[]{\includegraphics[width=0.3\linewidth]{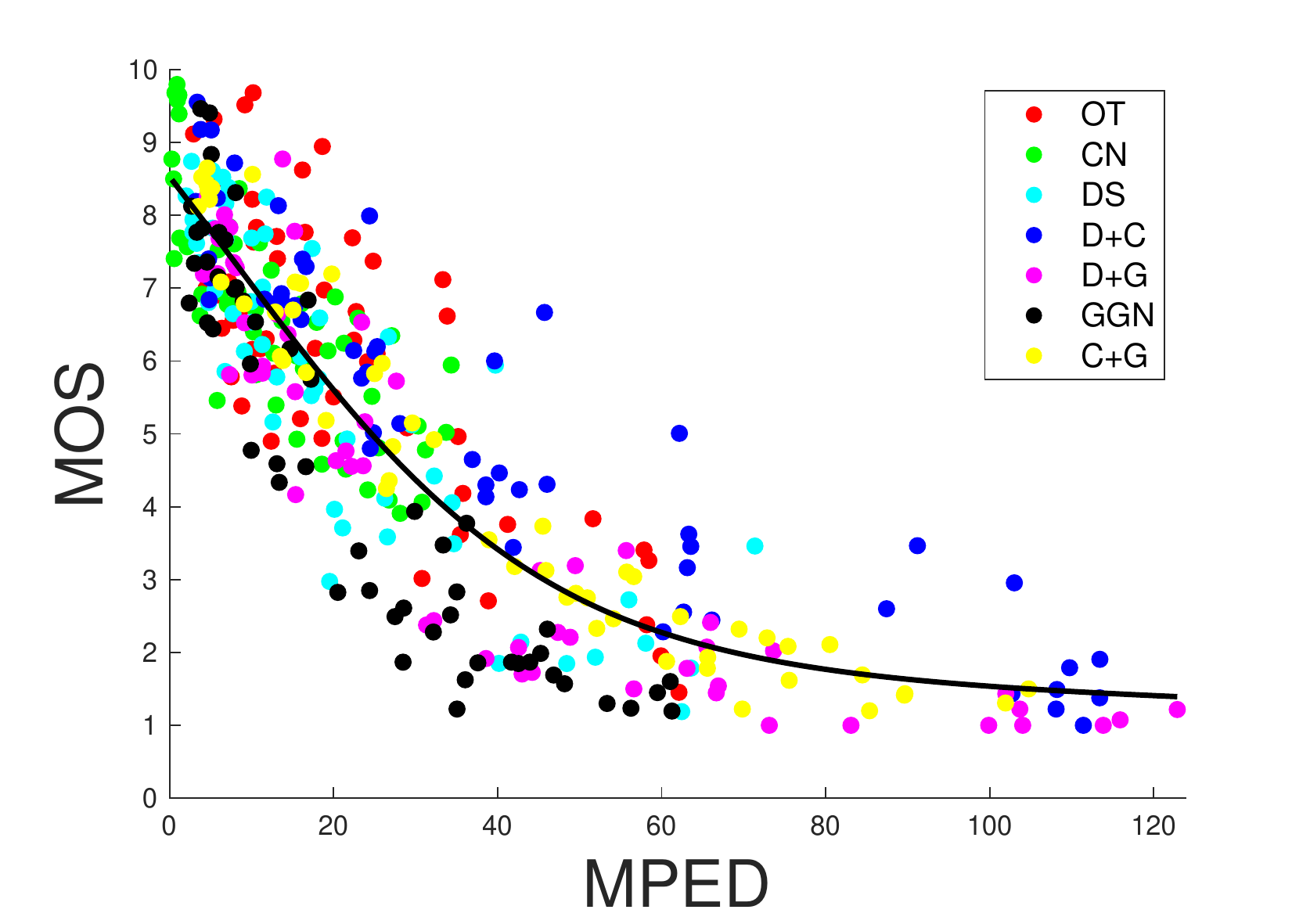}}
     \subfigure[]{\includegraphics[width=0.3\linewidth]{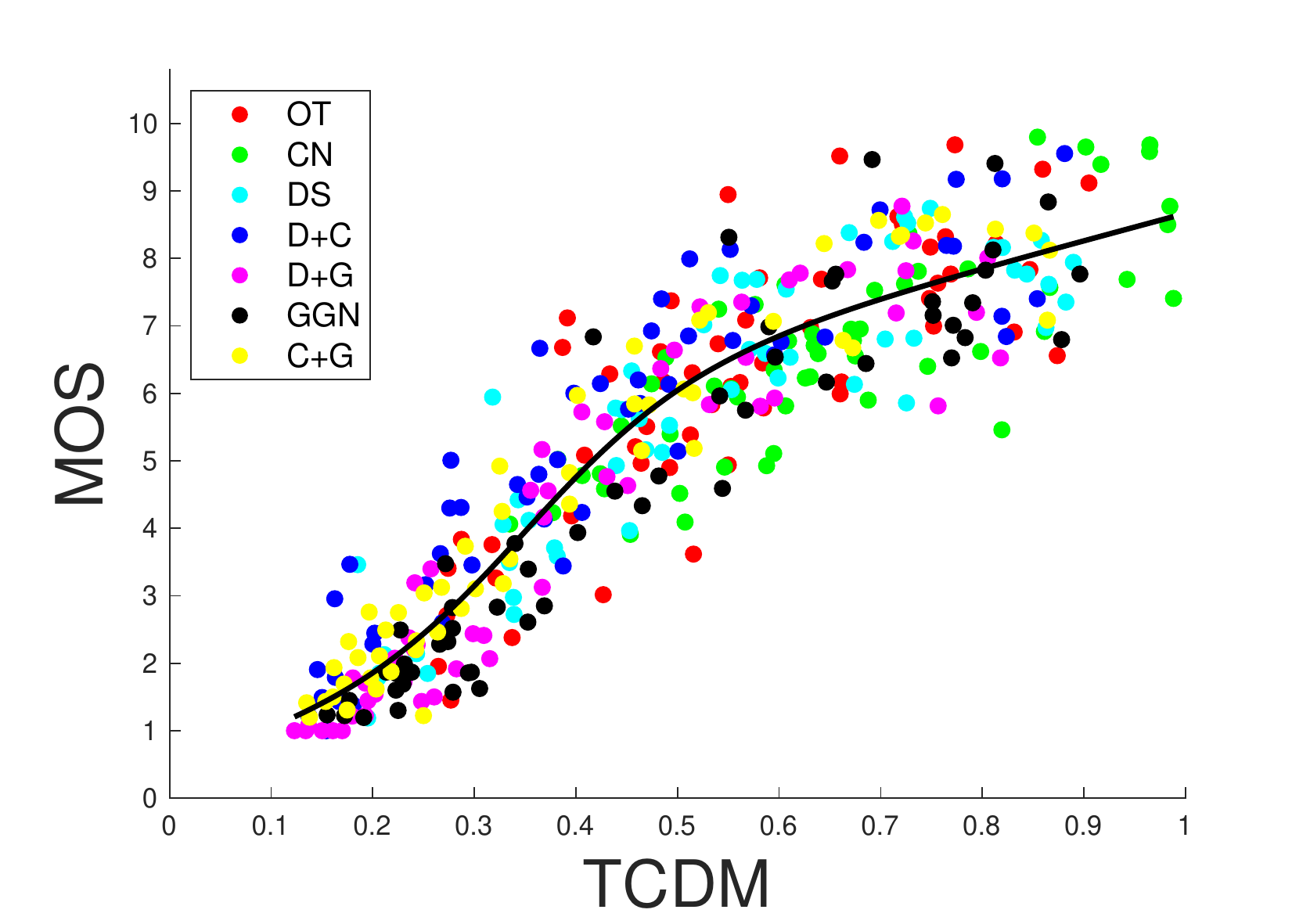}}
    \caption{Scatter plots of MOS values versus objective scores obtained by different models for the SJTU-PCQA database. (a) PCQM; (b) GraphSIM; (c) MS-GraphSIM; (d) pointSSIM; (e) MPED; (f) TCDM.}
    \label{fig:scatter_plots}
\end{figure*}

\begin{figure}
    \centering
    \includegraphics[width=\linewidth]{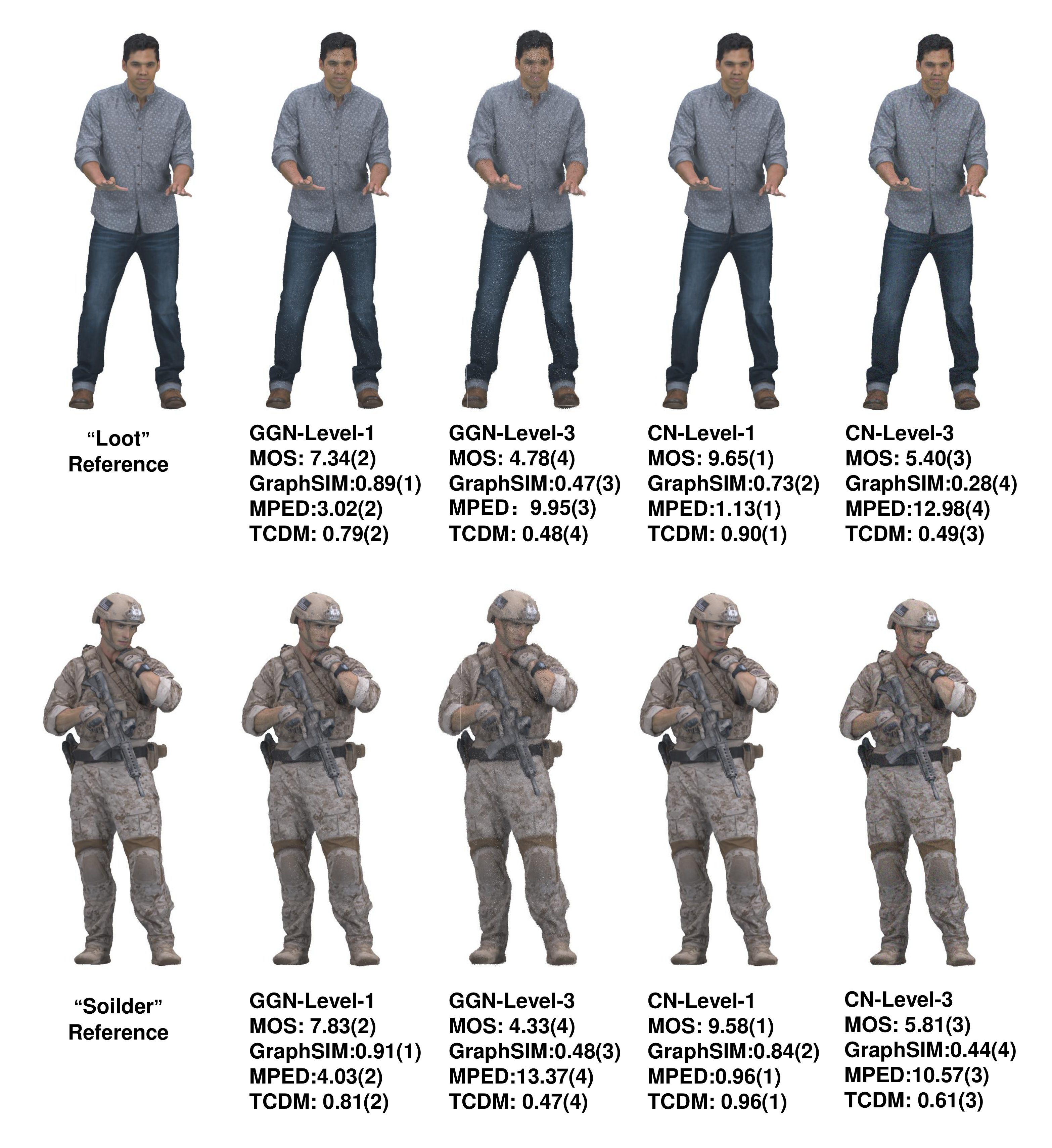}
  \caption{Example point clouds corrupted by geometry Gaussian noise and color noise. The associated MOS values and objective scores of three metrics (GraphSIM, MPED, TCDM) are also provided. The values of 1-4 represent the rank of subjective values and objective scores in four distorted samples generated from one reference. For instance, 7.34(2) under the "GGN-Level-1" sample of "Loot" indicates that the sample has the MOS value of 7.34, which ranks 2nd out of four distorted samples of "Loot".} 
    \label{fig:visual_sample}
\end{figure}

\section{Experimental Evaluations}\label{sec:experiment results}
This section evaluates the proposed method and other SOTA metrics for point cloud quality prediction, using five publicly accessible point cloud databases.
\subsection{Databases and Evaluation Protocols}
We review five PCQA databases used in our experiments as follows:

{$\bullet$ SJTU-PCQA database \cite{yang2020predicting}.}
There are 9 high-quality reference point cloud samples and 378 distorted samples provided with MOS. Each reference point cloud is disturbed by 7 different types of impairment under 6 levels, including four individual distortions, namely octree-based compression (OT), color noise (CN), geometry Gaussian noise (GGN), downsampling (DS), and three superimposed distortions, namely downsampling and color noise (D+C), downsampling and geometry Gaussian noise (D+G), color noise and geometry Gaussian noise (C+G).

{$\bullet$ WPC database \cite{liu2022perceptual}.}
There are 20 reference point clouds and 740 distorted point clouds generated from the references under 5 types of distortion, including downsampling, Gaussian noise contamination, geometry-based point cloud compression using the trisoup encoding module (G-PCC(T)) and the octree encoding module (G-PCC(O)), and video-based point cloud compression (V-PCC).

{$\bullet$ M-PCCD database \cite{alexiou2019comprehensive}.}
There are 8 reference point clouds and 232 distorted samples affected by MPEG encoders. The distortion results from the compression by V-PCC and four variations of G-PCC. The former is divided into five levels, and the latter is split into six levels. Note that M-PCCD provides the MOS values for both the reference and distorted samples. In our experiments, we only use the MOS values of the distorted samples.

{$\bullet$ ICIP2020 database \cite{perry2020quality}.} There are 6 reference point clouds and 90 distorted point clouds and all references and distorted samples are provided with MOS values. The distortion results from the compression algorithms, including V-PCC, G-PCC(T), and G-PCC(O) under 5 levels. Like M-PCCD, we only use the MOS values of the distorted samples.

{$\bullet$ IRPC database \cite{javaheri2020point}.} There are 6 reference point clouds and 54 distorted point clouds. Each native sample is distorted using three different compression methods (i.e., OT, G-PCC and V-PCC) with three compression levels (i.e., high-quality (HQ), medium-quality (MQ), and low-quality (LQ)). Note that IPRC provides three types of MOS caused by different rendering solutions. In our experiments, we use the MOS values resulting from the "RColor" rendering. Please refer to \cite{javaheri2020point} for more details.

To ensure consistency between subjective values and objective predictions from various models, we first map objective predictions from different models to the same dynamic range following the recommendations suggested by the video quality experts group (VQEG) \cite{video2003final}. Then, the Pearson linear correlation coefficient (PLCC), the Spearman rank order correlation coefficient (SROCC), and the root mean square error (RMSE) are utilized to evaluate the performance of different models, which indicate the linearity, monotonicity and accuracy, respectively. The larger PLCC or SROCC comes with better model performance. In contrast, the lower RMSE is better. To map the dynamic range of the scores from the objective quality assessment models in a common scale, the logistic regression recommended by VQEG is used. The formula is described below, according to \cite{video2003final},

\begin{equation}\label{eq:logic_fit}
    R_i = \beta_1(\frac{1}{2}-\frac{1}{1+\exp{\beta_2(Q_i-\beta_3)}})+\beta_4Q_i+\beta_5,
\end{equation}
where $Q_i$ is the quality score of the $i$-th point cloud calculated by the PCQA model, $R_i$ is the objective score after the regression, and $\beta_1$, $\beta_2$, $\beta_3$, $\beta_4$, and $\beta_5$ are the parameters fitted by minimizing the sum of squared errors between the regressive scores $R_i$ and the subjective scores.

\subsection{Parameter Setting}\label{sec:parameter_setting}
Several parameters are required to be determined in the proposed model:

\textit{$L$ in the space segmentation.}
In the space segmentation, we need to determine the number of generating seeds $L$ to create the 3D Voronoi diagram. We choose $L=400$ to balance effectiveness and complexity.

\textit{$K$ in the SA-VAR model.}
$K$ is used as the order of the SA-VAR model and as the number of neighbors in computing the point-wise difference for prediction terms. Large $K$ can effectively reduce the code length in the process of predictive coding while resulting in higher model complexity and computational cost. $K=20$ is finally set to achieve the trade-off between effectiveness and complexity.

\textit{$T$ in the similarity pooling.} We simply set $T=0.000001$.

\textit{$\alpha$ in the feature fusion.}
$\alpha$ used in Eq. \eqref{eq:feature_fusion} determines the relative weight of two quality features. The SJTU-PCQA database is utilized as the training pool to choose the optimal weighted coefficients. $\alpha = 0.3$ is finally determined. 

\begin{table*}[pt]
  \centering
  \caption{PERFORMANCE COMPARISON FOR FR-PCQA METRICS ON EASH INDIVIDUAL DISTORTION TYPE IN TERMS OF SROCC}
\resizebox{\textwidth}{!}{
\begin{tabular}{c|cccccccccc|c}
     \hline
    SJTU-PCQA   & {M-p2po} & {M-p2pl} & {H-p2po} & {H-p2pl} & {$\rm PSNR_{YUV}$} & {PCQM} & {GraphSIM} & {MS-GraphSIM} & {PointSSIM} & {MPED} & {TCDM} \\
    \hline
     OT &\bf{0.825}	&0.726	&\bf{0.849}	&0.774	&0.357	&0.758	&0.693	&0.714	&0.756 &0.679	&0.793 \\
   
     CN &-	&-	&-	&-	&0.753	&\bf{0.842}	&0.778	&0.770	&0.797 &\bf{0.824}	&{0.819}
     \\
   
    DS   &0.812	&0.62	&0.478	&0.521	&0.542	&0.808	&{0.872}	&0.864	&0.816 &\bf{0.878}	&\bf{0.876} \\

    GGN  &\bf{0.950}	&\bf{0.950}	&0.936	&0.947	&0.675	&0.905	&0.916	&0.916	&0.916 &0.925	&0.921
     \\
    
    D+C  &0.885	&0.506	&0.566	&0.533	&0.862	&{0.922}	&0.886	&0.914	&0.842 &\bf{0.937}	&\bf{0.934}
    \\
    
    D+G   &0.934	&0.925	&0.926	&\bf{0.946}	&0.61	&0.882	&0.888	&0.905	&0.913 &0.928	&\bf{0.944}
    \\
   
    C+G &0.951	&\bf{0.959}	&0.950	&{0.956}	&0.852	&0.922	&0.941	&0.951	&0.851 &\bf{0.969}	&0.951 \\

    \hline
       \multicolumn{1}{r}{} & \multicolumn{1}{r}{} & \multicolumn{1}{r}{} & \multicolumn{1}{r}{} & \multicolumn{1}{r}{} & \multicolumn{1}{r}{} & \multicolumn{1}{r}{} & \multicolumn{1}{r}{} & \multicolumn{1}{r}{} & \multicolumn{1}{r}{} & \multicolumn{1}{r}{} \\
    
     \hline
    WPC   & {M-p2po} & {M-p2pl} & {H-p2po} & {H-p2pl} & {$\rm PSNR_{YUV}$} & {PCQM} & {GraphSIM} & {MS-GraphSIM} & {PointSSIM} &MPED &{TCDM} \\
    \hline
    DS &\bf{0.901}	&\bf{0.905}	&0.849	&0.857	&0.707	&0.875	&0.898	&0.887	&0.836 &0.897	&0.882 \\

    C+G &0.729	&0.689	&0.738	&0.692	&0.777	&\bf{0.886}	&0.840	&{0.869}	&0.587 &\bf{0.880}	&{0.857} \\
    
    G-PCC(O)  &-	&-	&-	&-	&0.826	&\bf{0.894}	&{0.855}	&{0.859}	&0.792	&\bf{0.869}  &0.795 \\
   
    G-PCC(T) &0.465	&0.293	&0.463	&0.353	&0.646	&0.821	&0.816	&\bf{0.839}	&0.681 &0.551	&\bf{0.832} \\
  
    V-PCC  &\bf{0.698}	&0.445	&\bf{0.705}	&0.566	&0.345	&0.643	&0.612	&0.631	&0.366 &0.476	&0.640 \\
    \hline

    \multicolumn{1}{r}{} & \multicolumn{1}{r}{} & \multicolumn{1}{r}{} & \multicolumn{1}{r}{} & \multicolumn{1}{r}{} & \multicolumn{1}{r}{} & \multicolumn{1}{r}{} & \multicolumn{1}{r}{} & \multicolumn{1}{r}{} & \multicolumn{1}{r}{} & \multicolumn{1}{r}{} \\
    
     \hline
    M-PCCD   & {M-p2po} & {M-p2pl} & {H-p2po} & {H-p2pl} & {$\rm PSNR_{YUV}$} & {PCQM} & {GraphSIM} & {MS-GraphSIM} & {PointSSIM} &MPED &{TCDM} \\
    \hline
    G-PCC(O)-lifting	&0.900	&0.867	&0.899	&0.878	&0.722	&0.932	&0.963	&\bf{0.947}	&0.940 &0.904	&\bf{0.975}	\\
    G-PCC(O)-RAHT	&0.884	&0.842	&0.879	&0.851	&0.735	&0.923	&\bf{0.952}	&0.939	&0.940 &0.901 &\bf{0.969}	\\
    G-PCC(T)-lifting	&0.761	&0.466	&0.822	&0.690	&0.659	&0.893	&\bf{0.938}	&\bf{0.917}	&{0.900} &0.834	&0.898	\\
    G-PCC(T)-RAHT	&0.708	&0.397	&0.767	&0.613	&0.622	&0.895	&\bf{0.927}	&0.898	&\bf{0.958} &0.798	&0.899	\\
    V-PCC	&0.277	&0.149	&0.550	&0.125	&0.317	&0.736	&\bf{0.869}	&0.824	&\bf{0.845} &0.531	&0.814	\\

    \hline
    
    \multicolumn{1}{r}{} & \multicolumn{1}{r}{} & \multicolumn{1}{r}{} & \multicolumn{1}{r}{} & \multicolumn{1}{r}{} & \multicolumn{1}{r}{} & \multicolumn{1}{r}{} & \multicolumn{1}{r}{} & \multicolumn{1}{r}{} & \multicolumn{1}{r}{} & \multicolumn{1}{r}{} \\
    
     \hline
    ICIP2020   & {M-p2po} & {M-p2pl} & {H-p2po} & {H-p2pl} & {$\rm PSNR_{YUV}$} & {PCQM} & {GraphSIM} & {MS-GraphSIM} & {PointSSIM} &MPED &{TCDM} \\
    \hline
    G-PCC(O)	&0.904	&0.915	&\bf{0.942}	&0.908	&0.885	&\bf{0.968}	&0.831	&0.855	&0.812 &0.899	&0.885	\\
    G-PCC(T)	&0.908	&0.862	&0.892	&0.903	&0.817	&{0.954}	&0.941	&0.939	&0.910 &\bf{0.955}	&\bf{0.970}	\\
    V-PCC	&0.723	&0.295	&0.828	&0.708	&{0.848}	&\bf{0.960}	&0.720	&0.770	&0.822 &\bf{0.904}	&0.822	\\

    \hline

    \multicolumn{1}{r}{} & \multicolumn{1}{r}{} & \multicolumn{1}{r}{} & \multicolumn{1}{r}{} & \multicolumn{1}{r}{} & \multicolumn{1}{r}{} & \multicolumn{1}{r}{} & \multicolumn{1}{r}{} & \multicolumn{1}{r}{} & \multicolumn{1}{r}{} & \multicolumn{1}{r}{} \\
    
     \hline
    IRPC   & {M-p2po} & {M-p2pl} & {H-p2po} & {H-p2pl} & {$\rm PSNR_{YUV}$} & {PCQM} & {GraphSIM} & {MS-GraphSIM} & {PointSSIM} &MPED &{TCDM} \\
    \hline
    OT	&0.818	&0.892	&0.895	&0.902	&0.538	&0.868	&0.858	&0.849	&{0.912} &\bf{0.925}	&\bf{0.913}	\\
    G-PCC(T)	&0.656	&0.629	&0.525	&0.898	&0.831	&{0.901}	&0.792	&0.809	&0.588 &\bf{0.916}	&\bf{0.907}	\\
    V-PCC	&0.186	&0.282	&0.254	&\bf{0.435}	&\bf{0.686}	&0.554	&0.230	&0.272	&0.329 &0.415	&0.336	\\

    \hline
    \end{tabular}}
  \label{tab:test for distortionl}%
\end{table*}%

\setlength{\tabcolsep}{2.5pt}

\begin{figure*}
    \centering
    \subfigure[]{\includegraphics[width=0.32\linewidth]{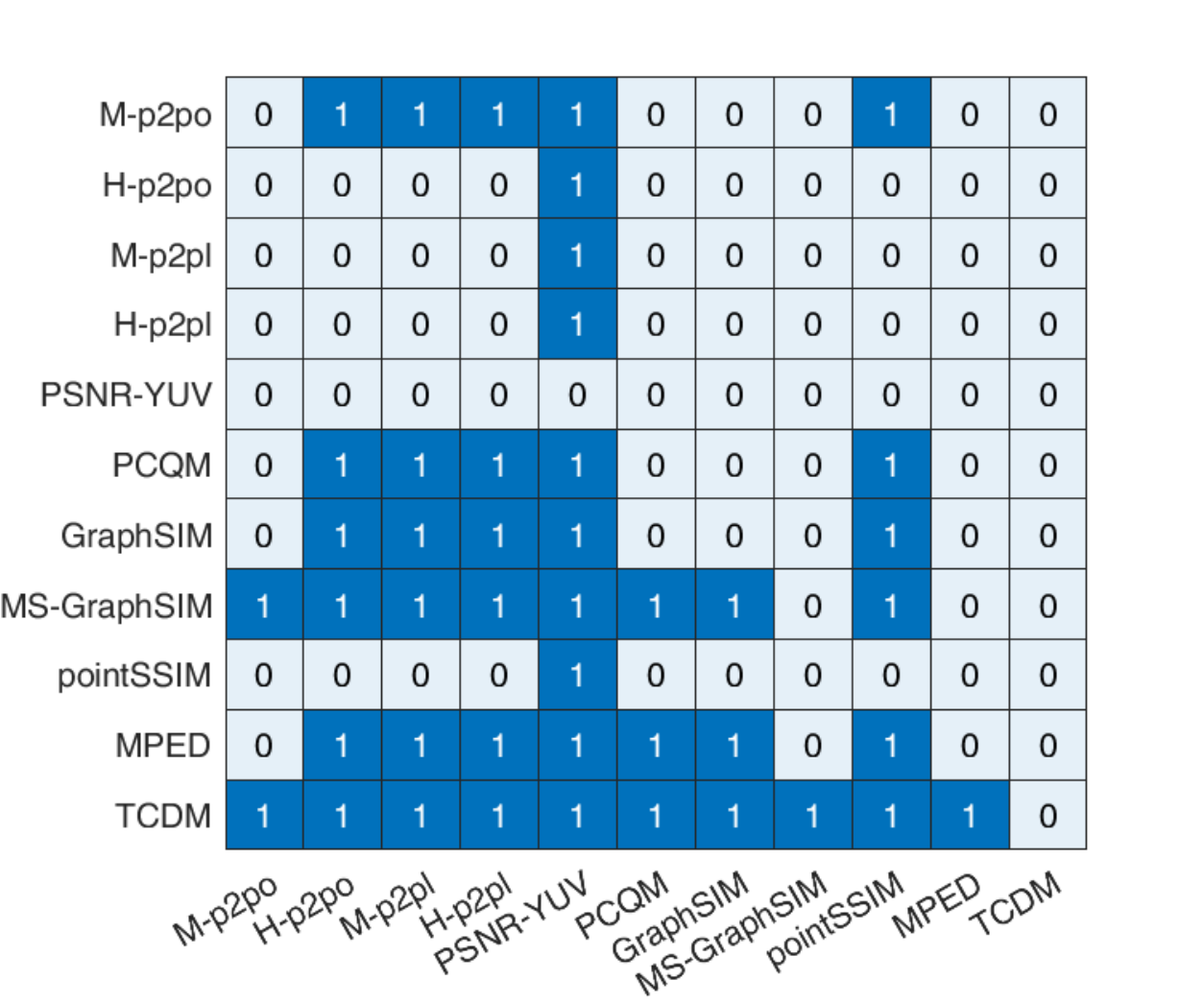}}
    \subfigure[]{\includegraphics[width=0.32\linewidth]{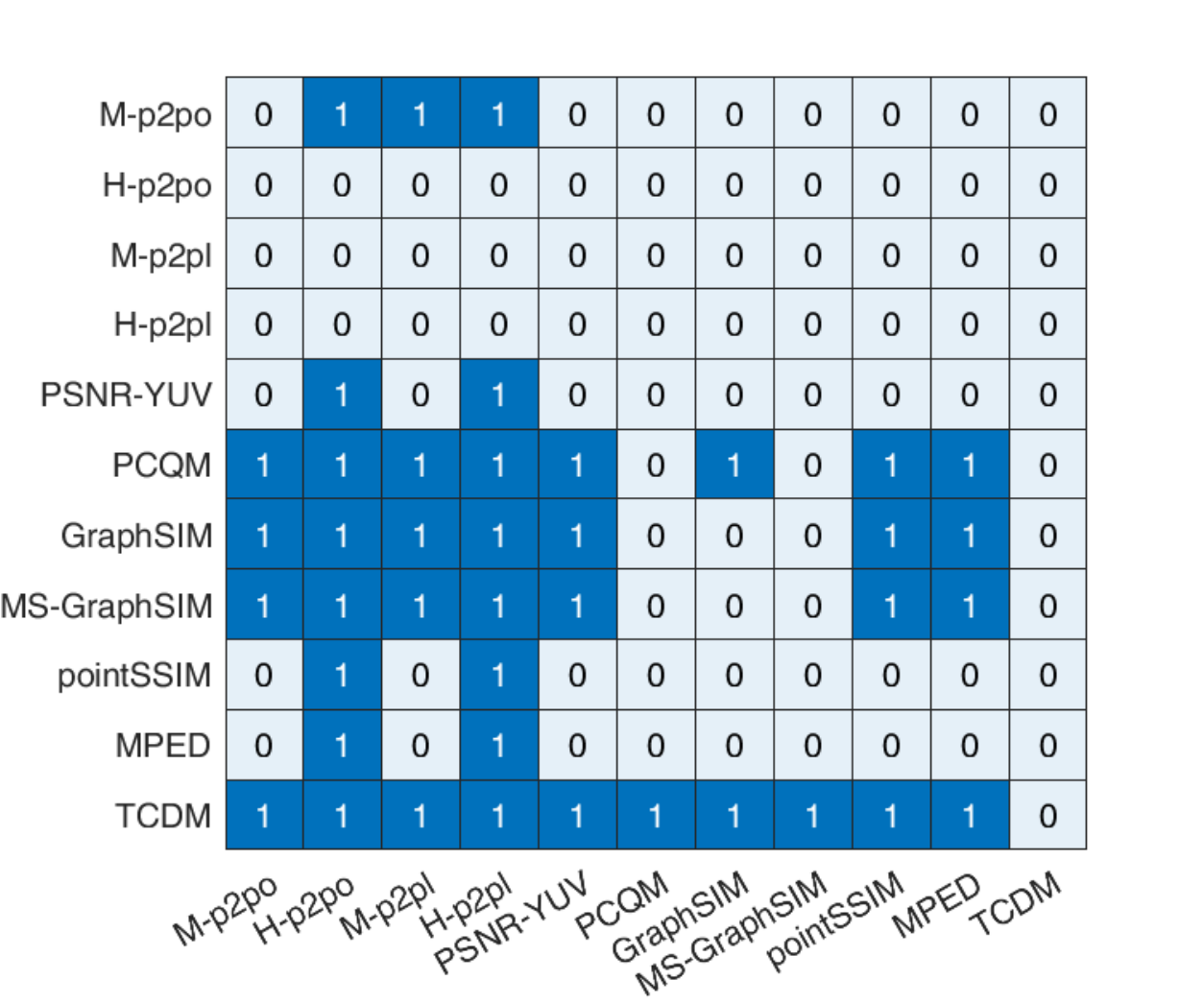}}
    \subfigure[]{\includegraphics[width=0.32\linewidth]{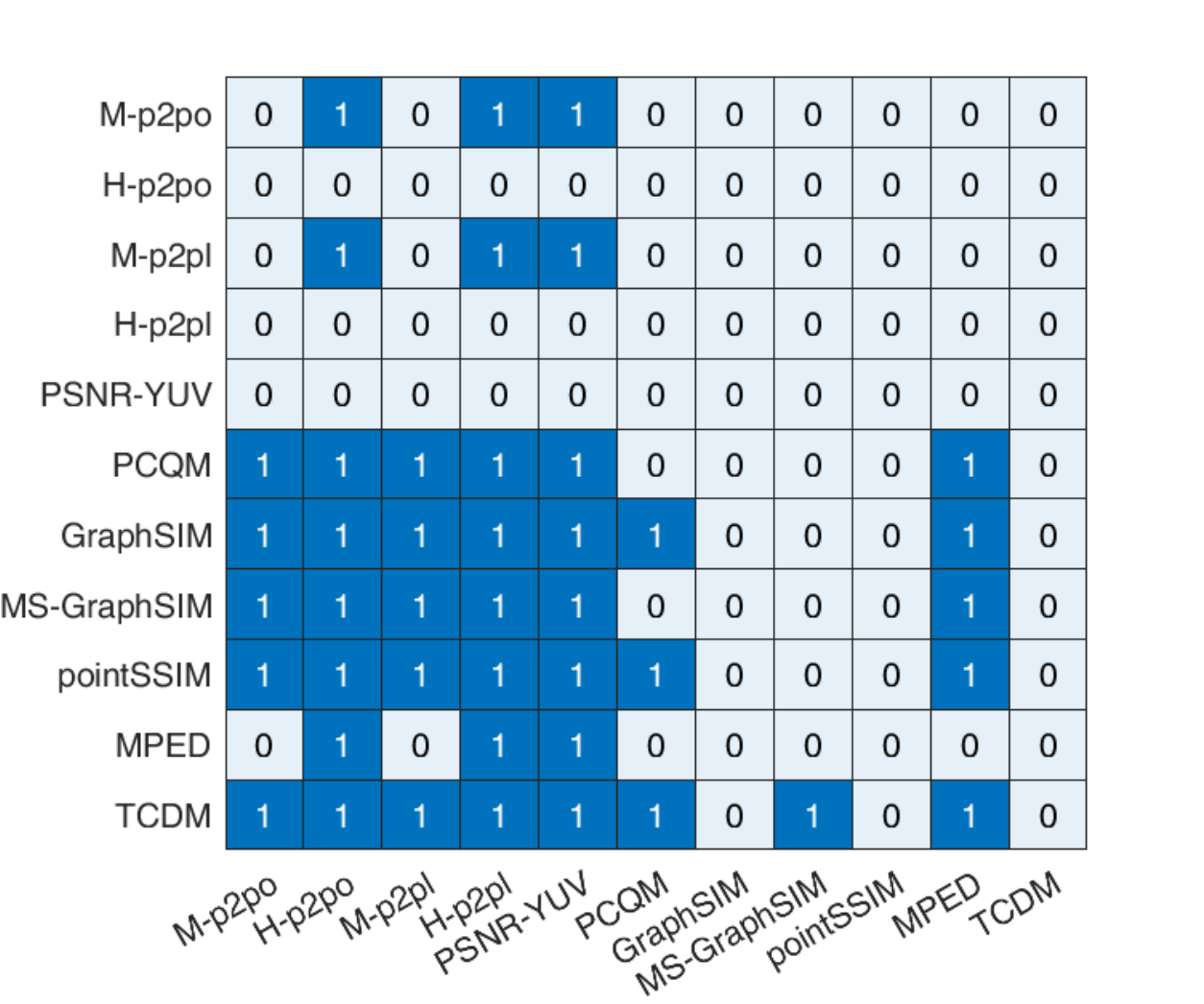}}
    \subfigure[]{\includegraphics[width=0.32\linewidth]{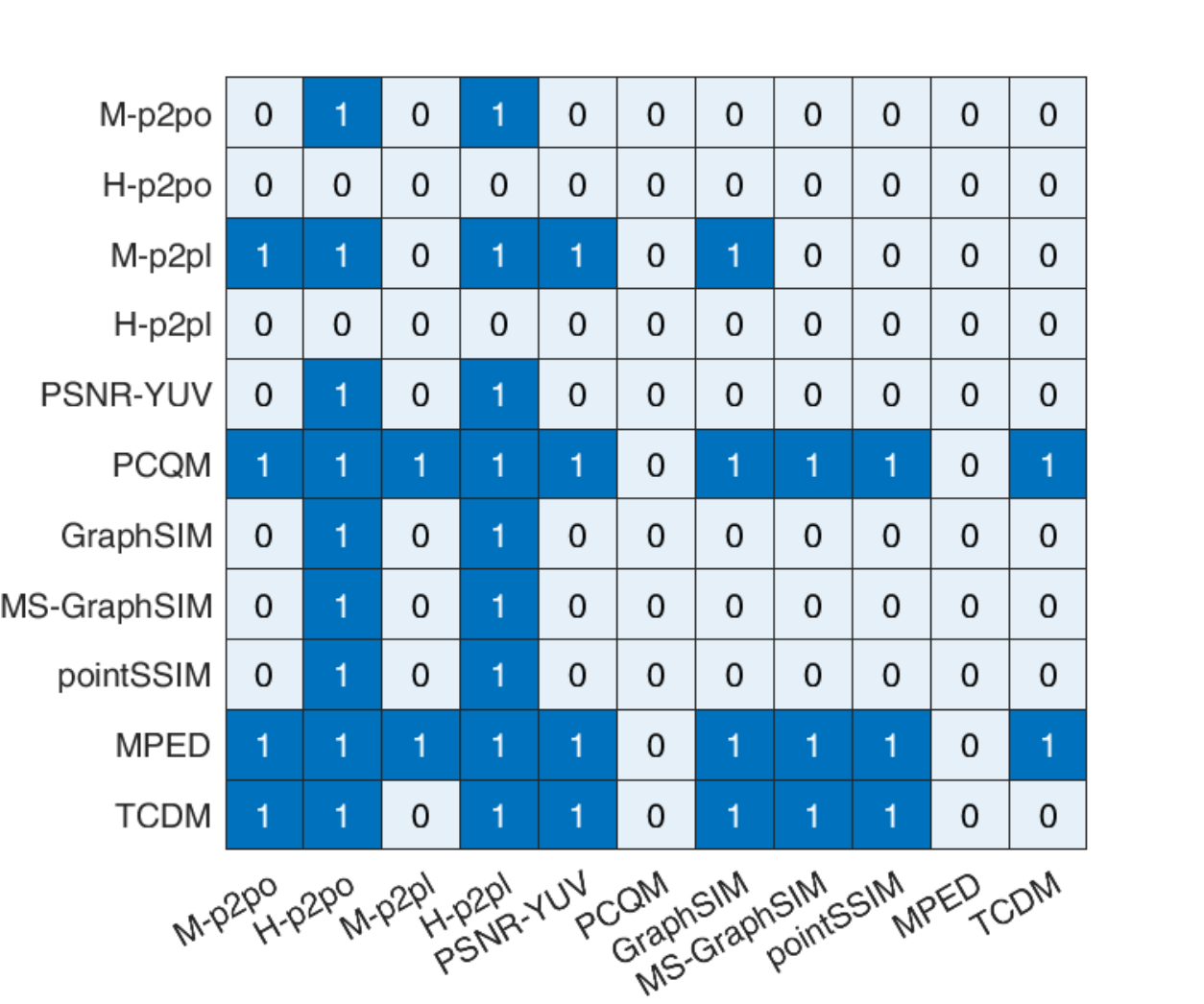}}
     \subfigure[]{\includegraphics[width=0.32\linewidth]{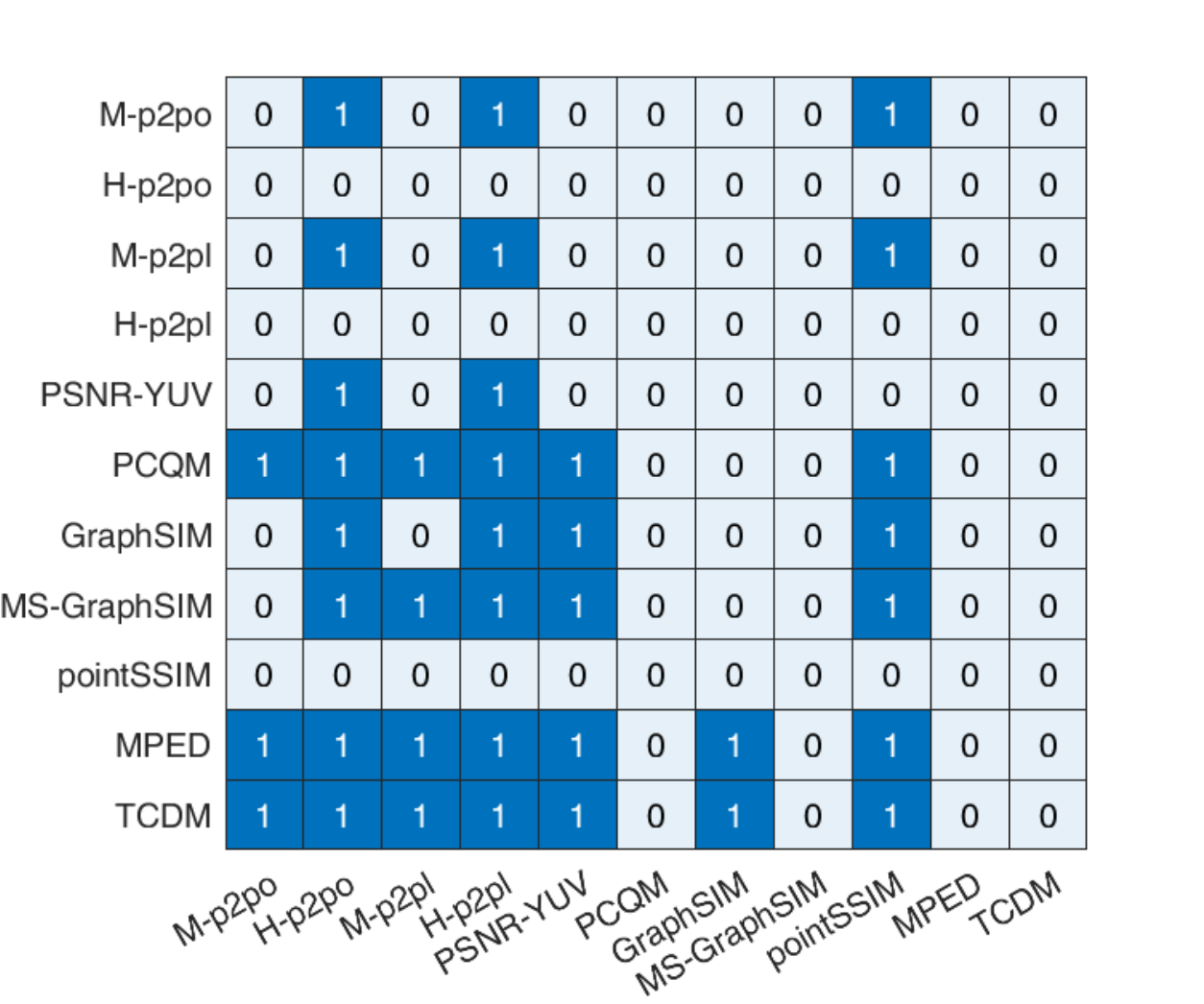}}
  \caption{The results of statistical significance tests of the competing PCQA models on (a) SJTU-PCQA; (b) WPC; (c) M-PCCD; (d) ICIP2020; (e) IRPC. A value of ‘1’ (highlighted in navy blue) indicates that the model in the row is significantly better than the model in the column, while a value of ‘0’ (highlighted in light blue) indicates that the first model is not significantly better than the second one.} 
    \label{fig:sigficance_test}
\end{figure*}

\subsection{Overall Performance Comparison}
Table \ref{tab:overall_performance} lists the experimental results on the five PCQA databases in terms of SROCC, PLCC, and RMSE. The proposed method, TCDM, is compared with several FR-PCQA metrics, and we run all published codes to produce the results (note that the first five point-based metrics are reproduced by Matlab following the implementation of \cite{MPEGSoft}). We use the PSNR measure for the first five point-based metrics and the L1 norm for MPED \cite{yang2021mped} and the color measure for pointSSIM \cite{alexiou2020towards}.

\begin{table*}[pt]
  \centering
  \caption{PERFORMANCE COMPARISON FOR FR-PCQA METRICS ON SIAT-PCQD and WPC2.0 databases}

\begin{tabular}{c|cccccccccc|c}

    \multicolumn{1}{r}{} & \multicolumn{1}{r}{} & \multicolumn{1}{r}{} & \multicolumn{1}{r}{} & \multicolumn{1}{r}{} & \multicolumn{1}{r}{} & \multicolumn{1}{r}{} & \multicolumn{1}{r}{} & \multicolumn{1}{r}{} & \multicolumn{1}{r}{} & \multicolumn{1}{r}{} \\
    
     \hline
    SIAT-PCQD   & {M-p2po} & {M-p2pl} & {H-p2po} & {H-p2pl} & {$\rm PSNR_{YUV}$} & {PCQM} & {GraphSIM} & {MS-GraphSIM} & {PointSSIM} &MPED &{TCDM} \\
    \hline
    PLCC &0.213	&0.219	&0.182	&0.201	&0.345	&0.497	&0.700	&0.728	&0.700	&0.525	&\textbf{0.736}
	\\
    SROCC 	&0.170	&0.185	&0.179	&0.185	&0.339	&0.618	&0.613	&0.628	&0.631	&0.473	&\textbf{0.658}
	\\
    RMSE  &0.129	&0.128	&0.129	&0.129	&0.123	&0.131	&0.094	&0.090	&0.094	&0.112	&\textbf{0.089}
	\\

    \hline

    \multicolumn{1}{r}{} & \multicolumn{1}{r}{} & \multicolumn{1}{r}{} & \multicolumn{1}{r}{} & \multicolumn{1}{r}{} & \multicolumn{1}{r}{} & \multicolumn{1}{r}{} & \multicolumn{1}{r}{} & \multicolumn{1}{r}{} & \multicolumn{1}{r}{} & \multicolumn{1}{r}{} \\
    
     \hline
    WPC2.0   & {M-p2po} & {M-p2pl} & {H-p2po} & {H-p2pl} & {$\rm PSNR_{YUV}$} & {PCQM} & {GraphSIM} & {MS-GraphSIM} & {PointSSIM} &MPED &{TCDM} \\
    \hline
    PLCC	&0.620	&0.632	&0.44	&0.539	&0.458	&0.726	&0.745	&0.773	&0.570	&0.623	&\textbf{0.803}	\\
    SROCC 	&0.583	&0.594	&0.350	&0.400	&0.452	&0.726	&0.751	&0.777	&0.568	&0.628	&\textbf{0.802}	\\
    RMSE    &17.260	&17.048	&19.757	&18.533	&19.557	&15.133	&14.668	&13.963	&18.083	&17.206	&\textbf{13.100}
	\\

    \hline
    \end{tabular}
  \label{tab:test for VPCC}%
\end{table*}%

\begin{figure}
    \centering
    \includegraphics[width=\linewidth]{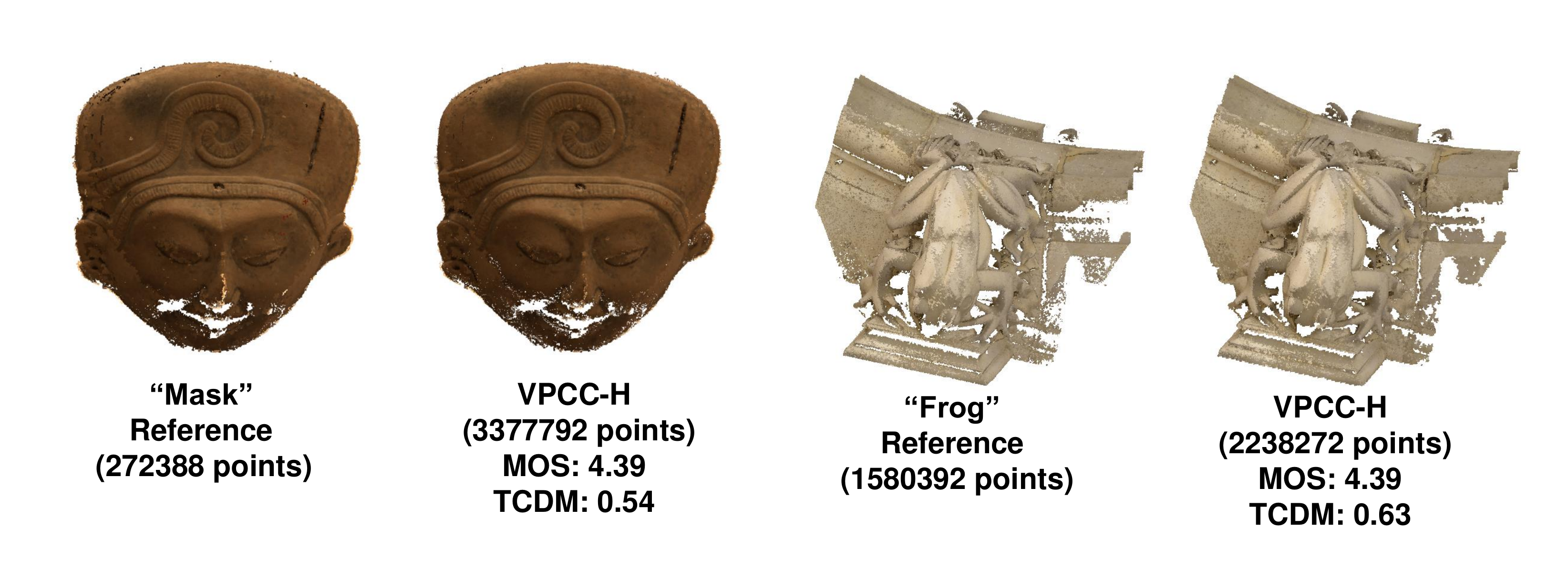}
    \caption{Example point clouds corrupted by V-PCC. The associated MOS values and TCDM scores are provided for each distorted samples. The point numbers of both reference and distorted samples are presented under each sample. }
    \label{fig:IRPC-VPCC}
\end{figure}

\znote{For each database, the top two results for each evaluation criterion are highlighted in \textbf{boldface}. We can see that the proposed TCDM is among the top two metrics on four databases except for ICIP 2020, and provides the best PLCC and RMSE on three databases including SJTU-PCQA, WPC and M-PCCD. Note that SJTU-PCQA, WPC and M-PCCD have more samples than ICIP2020 and IRPC. Thus, the evaluation results on these three databases may be more convincing. Although some models provide outstanding results in certain instances (e.g., PCQM on ICIP2020 and MPED on IRPC), they perform less well in other cases (e.g., PCQM and MPED on M-PCCD). Furthermore, to measure the average performance of these PCQA models across multiple databases, we derive the average ranking of each method with respect to competing methods in terms of PLCC and SROCC (see the last two columns of Table \ref{tab:overall_performance}). It is evident that our method outperforms other PCQA metrics with a higher average ranking of PLCC and SROCC at (1.6, 1.8), followed by (3.2, 3.0) from PCQM and (3.2, 3.2) from MPED. }

For better illustration, we also provide the scatter plots shown in Fig. \ref{fig:scatter_plots} for multiple competing PCQA metrics (PCQM, GraphSIM, MS-GraphSIM, pointSSIM, MPED, TCDM) on the SJTU-PCQA database. In all the plots, each point represents a test point cloud. The vertical axis denotes the subjective rating produced by human observers and the horizontal axis denotes the metric score produced by objective metrics. 
The black curves in the scatter plots are obtained by the nonlinear fitting in Eq. \eqref{eq:logic_fit}. 
The algorithm's performance is considered to be superior when the scattered points are positioned in close proximity to the fitted curve. According to Fig. \ref{fig:scatter_plots}, it can be clearly seen that the objective scores predicted by TCDM correlate better with human judgements than other competitors because more points converge to the fitted curve, which further demonstrates the effectiveness of our method. Another interesting observation is that the scattered points of TCDM correlate better in the low-quality range than in the high-quality range. The reason may be that TCDM is more capable of capturing these "visible differences" while samples in the high-quality range include many invisible distortions due to the masking effect. One possible solution is to apply different strategies for high-quality samples and low-quality samples such as \cite{larson2010most}. 

We further present some illustrative examples of distorted samples (from the SJTU-PCQA database) in Fig. \ref{fig:visual_sample}, where the corresponding distortion levels, MOS values, and objective scores of three metrics (GraphSIM, MPED, TCDM) are provided. We also annotate the rank of the MOS values and three types of objective scores for each four distorted samples generated from the same reference. For instance, "7.34(2)" under the "GGN-Level-1" sample of "Loot" means that the sample has the MOS value of 7.34, and it ranks 2nd out of four distorted samples of "Loot". Note that higher GraphSIM and TCDM scores and lower MPED scores indicate better quality. According to Fig. \ref{fig:visual_sample}, we can see that: i) for each distortion type, as the point clouds are arranged in decreasing order of their MOS values, the objective scores of the three metrics are in the same order as well, which states that all three metrics can accurately capture the rank among different levels of the same distortion; ii) GraphSIM and MPED both fail to reflect the correct rank between two distortion types. For example, the distorted sample of "Loot" under the "CN-Level-3" scheme has a higher MOS than that under the "GGN-Level-3" scheme, but the objective scores provided by GraphSIM and MPED present opposite rank. The correct subjective rank for four distorted samples of "Loot" is "2-4-1-3", but the objective ranks of GraphSIM and MPED are "1-3-2-4" and "2-3-1-4", respectively. In comparison, TCDM captures the difference between the two types of distortion and presents lower scores for samples affected by GGN, which contributes to the correct objective rank. Based on the above example, we can conclude that TCDM is more capable of differentiating the quality among various types of distortion, leading to superior overall performance.

In order to draw statistically meaningful conclusions regarding the model performance, we conduct a series of hypothesis tests using the prediction residuals obtained from each model's nonlinear regression. The findings of these significance tests are depicted in Fig. \ref{fig:sigficance_test}. Assuming that the prediction residuals of the models adhere to a Gaussian distribution, we employed the left-tailed F-test to compare the residuals of every pair of models under examination. With a significance level of 0.05, a left-tailed F-test yielding a value of $H=1$ indicates that the first model (as denoted by the corresponding row in Fig. \ref{fig:sigficance_test}) exhibits better performance than the second model (as denoted by the respective column in Fig. \ref{fig:sigficance_test}) with confidence exceeding $95\%$. Conversely, a value of $H=0$ suggests that the first model does not possess a statistically significant advantage over the second model. \znote{In total, TCDM achieves the value of "1" 42 times, followed by PCQM (34 times), MS-GraphSIM (28 times) and MPED (28 times). This shows that TCDM is superior to other metrics. We can see that TCDM performs best on the SJTU-PCQA, WPC, and M-PCCD databases. On the ICIP2020 database, TCDM is significantly better than all other models, except for PCQM and MPED. On the IRPC database, TCDM and MPED both perform well and their performance does not show statistically significant difference.}

\begin{table}[t]
	\caption{ \MakeUppercase{Performance comparison OF DIFFERENT color spaces on SJTU-PCQA}} \label{tab:color_space}
	\centering
	\begin{tabular}{|c|c|c|c|}
		\hline
		Color Space & PLCC & SROCC & RMSE \\ \cline{1-4}
		RGB & 0.9301 & 0.9102 & 0.8912 \\ \cline{1-4}
		YUV & 0.9294 & 0.9096 & 0.8957 \\ \cline{1-4}
		GCM & 0.9289 & 0.9099 & 0.8987 \\ \cline{1-4}
	\end{tabular}
\end{table}

\subsection{Performance Comparison on Individual Distortion Type}
Good (bad) overall performance does not necessarily mean
good (bad) performance for individual distortion types. Therefore, we compare the performance of the FR-PCQA metrics towards different types of distortion on the five databases. SROCC is used as the only evaluation measure and the results are shown in Table \ref{tab:test for distortionl}. For each type of distortion in each database, we also use \textbf{boldface} to highlight the algorithm with the top two SROCCs among all competing metrics. 

\begin{table}[pt]
  \centering
  \caption{ \MakeUppercase{Performance comparison OF DIFFERENT Segmentation resolutions on SJTU-PCQA} }
    \begin{tabular}{|c|c|c|c|c|}
    \hline
     Strategy   & {$L$} & {PLCC } & {SROCC} & {RMSE} \\
    \hline
   
    \multirow{4}{*}{RS} &100	&0.9282	&0.9079	&0.903	\\ \cline{2-5}
&200	&\bf{0.9313}	&\bf{0.9118}	&\bf{0.8841}	\\ \cline{2-5}
&400	&0.9293	&0.9102	&0.8961	\\ \cline{2-5}
&1000 &0.9270	&0.9099	&0.9102\\
\hline
\hline
    \multirow{4}{*}{FPS} 
&100	&0.9291	&0.9075	&0.8976	\\ \cline{2-5}
&200	&\bf{0.9306}	&{0.9099}	&\bf{0.8881}	\\ \cline{2-5}
&400	&0.9301	&\bf{0.9102}	&0.8912	\\ \cline{2-5}
&1000	&0.9260	&0.9080	&0.9161
\\
    \hline
\hline
 {\textit{w/o} VD} &1 &0.9240 &0.9009 &0.9279 \\
    \hline
    
    \end{tabular}%
  \label{tab:ablation_sampling}%
\end{table}%

There are 23 types of distortion on the five databases. According to Table \ref{tab:test for distortionl}, MPED is among the top two models 10 times, followed by 9 times for the proposed TCDM and 6 times for PCQM. In general, our method is more capable of coping with the distortions caused by downsampling and G-PCC compression. We notice that TCDM provides lower SROCC than PCQM and MPED over V-PCC on the ICIP2020 and IRPC databases, which leads to its relatively inferior performance on these databases. Another noticeable phenomenon is that TCDM shows inconsistent prediction performance of V-PCC across different databases. Specifically, for V-PCC, TCDM performs relatively well on M-PCCD while providing poor results on IRPC. Looking further into this issue, we find that the extent of the point number increase caused by V-PCC relies on the sparsity of the reference point cloud, which further influences the performance of TCDM. We show several visual samples (provided in the IRPC database) in Fig. \ref{fig:IRPC-VPCC} to illustrate the phenomenon. Given two references "Mask" with 272388 points and "Frog" with 1580392 points, their V-PCC compressed samples at the high-qualiy level (denoted as "VPCC-H") share the same MOS values of 4.39. However, the sparsity of the "Mask" reference results in a $1140\%$ increase of the point number (from 272388 to 3377792) after V-PCC, whereas the "VPCC-H" sample of the "Frog" reference only exhibits a $42\%$ increase (from 1580392 to 2238272). Given that TCDM needs to find $K$ nearest neighbors in the distorted point cloud for each point of the reference sample, the significant increase in the point number of "Mask" inevitably impacts the kNN search process of TCDM, consequently influencing the prediction of objective scores. Specifically, TCDM exhibits a lower objective score for the compressed sample of "Mask", thereby impairing the performance of TCDM on V-PCC.

It should be noted that the significant increase in the point number of "Mask" is not common, as V-PCC is designed to compress dense point clouds, such as the "Frog" sample. We further test TCDM and competing metrics on two PCQA databases that consist only of V-PCC distorted samples, i.e., the SIAT-PCQD database (340 samples) \cite{wu2021subjective} and the WPC2.0 database (400 samples) \cite{liu2021reduced} . Note that the reference point clouds used in the SIAT-PCQD and WPC2.0 databases are all dense. We show the performance of different metrics on the SIAT-PCQD and WPC2.0 databases in Table \ref{tab:test for VPCC}. According to the table, we can see that TCDM outperforms other competing metrics in terms of three criteria, which demonstrate the effectiveness of TCDM in most cases of V-PCC.

\begin{table}[pt]
  \centering
  \caption{\MakeUppercase{PERFORMANCE COMPARISON (IN TERMS OF SROCC) OF different spatial weights}}
  \resizebox{\linewidth}{!}{
    \begin{tabular}{|c|c|c|c|c|c|c|}
    \hline
    \multicolumn{2}{|c|}{Database} & {SJTU} & {WPC} & {M-PCCD} & {ICIP2020} & {IRPC}  \\
    \hline
    \multirow{5}{*}{$d_{i,j}$} & 1 &\bf{0.9143}	&0.8021	&\bf{0.9478}	&0.9300	&0.8051 
    \\
    \cline{2-7} 
    & ${1}/{\|\x_i-\x_j\|_2}$ &0.8937	&0.7975	&0.9349	&\bf{0.9367}	&0.7934 
    \\
    \cline{2-7} 
    & $\exp(-\|\x_i-\x_j\|_2/\eta)$ &0.9056	&0.7982	&0.9354	&0.9322	&0.8198 
    \\
    \cline{2-7} 
    & Proposed &0.9102	&\bf{0.8041}	&0.9437	&0.9354	&\bf{0.8401} 
  \\
    \hline
    \end{tabular}}%
  \label{tab:ablation_spatial_weight}%
\end{table}%

\subsection{Impact of Key Modules and Parameter Values}\label{sec:ablation_study}

\subsubsection{Color Space}
We have exemplified the effectiveness of TCDM based on the RGB-based color channel decomposition. In this part, we test the performance of TCDM in terms of other color channel spaces, e.g., YUV and Gaussian color model (GCM)\cite{geusebroek2000color}. YUV and GCM are two color channel spaces that consist of one luminance component and two chrominance components. Given that the HVS is more sensitive to the luminance component, 
we set $k_l=6$, $k_{c1}=k_{c2}=1$ following \cite{yang2020inferring} when calculating the point-wise difference vector in Eq. \eqref{eq:F2_step1}. $k_l$ represents the weighting factor of the luminance component and $k_{c1}$, $k_{c2}$ are for the chrominance components.
We use SJTU-PCQA as the test database and the results are shown in Table \ref{tab:color_space}. From the table, we observe that the results of TCDM are very close for multiple color spaces, which validates the robustness of TCDM.

\subsubsection{Segmentation Resolution}
In Section \ref{sec:segmentation}, we use the farthest point sampling (FPS) to obtain $L$ seeds that generate the 3D Voronoi diagram, where $L$ determines the number of Voronoi's divisions. We further test the model performance with random sampling (RS) at various $L$ values. As shown in Table \ref{tab:ablation_sampling}, we can see that both RS and FPS retain a consistent correlation with MOS values across different $L$. $L=400$ offers the best SROCC under FPS, while $L=200$ outperforms other segmentation resolutions under RS. It is also observed that a too large or too small $L$ degrades the performance slightly. Although FPS can attain more uniform sampling than RS, the 3D Voronoi diagram achieves non-overlapping and non-omission divisions for both two strategies. This indicates that our model can maintain reliable performance under different sampling strategies and segmentation resolutions.

\znote{Furthermore, we present the model performance without performing the space segmentation in the last row (denoted by \textit{w/o} VD) of Table \ref{tab:ablation_sampling}, which can be considered the global processing with $L=1$. We can see that global processing provides poorer performance compared to the local processing, which justifies the effectiveness of the space segmentation.}

\begin{table}[pt]
  \centering
  \caption{ PERFORMANCE COMPARISON (IN TERMS OF SROCC) OF DIFFERENT NEIGHBORHOOD SCALES}
    \begin{tabular}{|c|c|c|c|c|c|c|}
    \hline
    \multicolumn{2}{|c|}{Database} & \multicolumn{1}{c|}{SJTU-PCQA} & \multicolumn{1}{c|}{WPC} & \multicolumn{1}{c|}{M-PCCD} & \multicolumn{1}{c|}{ICIP2020} & \multicolumn{1}{c|}{IRPC} \\
    \hline
    \multirow{4}{*}{K} & 10    & 0.9009  & 0.7757  & 0.9379  & 0.9307  & 0.8293  \\
\cline{2-7}          & 20    & 0.9102  & \bf{0.8041}  & 0.9437  & \bf{0.9354}  & 0.8401  \\
\cline{2-7}          & 30    & 0.9154  & 0.7806  & 0.9454  & 0.9295  & 0.8485  \\
\cline{2-7}          & 50    & \bf{0.9159}  & 0.7695  & \bf{0.9456}  & 0.9275  & \bf{0.8543}  \\
    \hline
    \end{tabular}%
  \label{tab:ablation_neighbor_scale}%
\end{table}%

\subsubsection{Spatial Weight}
The spatial weight in Eq. \eqref{eq:SA-VAR} is employed to strengthen the AR’s capability of capturing
irregular spatial structure. We further test the model performance (in terms of SROCC) with different forms of $d_{i,j}$ in Eq. \eqref{eq:SA-VAR} and show the result in Table \ref{tab:ablation_spatial_weight}. We can see that not all distance-related spatial weights provide better performance than a constant weight. For instance, $d_{i,j}=1/\|\x_i-\x_j\|_2$ shows poorer correlation than $d_{i,j}=1$, which may be due to its wide range of $(0,+\infty)$ and steep function curve at small distances. In comparison, the proposed $d_{i,j}$ is in the range of [0.5,1] , which avoids overestimating or underestimating some points too much. In general, the proposed spatial weight presents better SROCC than $d_{i,j}=1$ on three databases (i.e, WPC, ICIP2020, and IRPC). In particular, the proposed weight yields a noticeable gain on IRPC over $d_{i,j}=1$, which states that introducing proper spatial weight into the AR model benefits PCQA.

\subsubsection{Neighborhood Scale}
$K$ is used as the order of the SA-VAR model and as the number of neighbors in computing point-wise differences for prediction terms. We test the performance of TCDM under different $K$ values to investigate the influence of the scales and report the results in Table \ref{tab:ablation_neighbor_scale}. We see that $K=20$ provides the best performance on WPC and ICIP2020 while $K=50$ stands out on the other three databases. It is worth noting that larger $K$ leads to higher computational cost. Therefore, we choose $K=20$ in our model to achieve a better trade-off between effectiveness and complexity.

\subsubsection{Weight Allocation}
\begin{figure}
    \centering
    \includegraphics[width=1\linewidth]{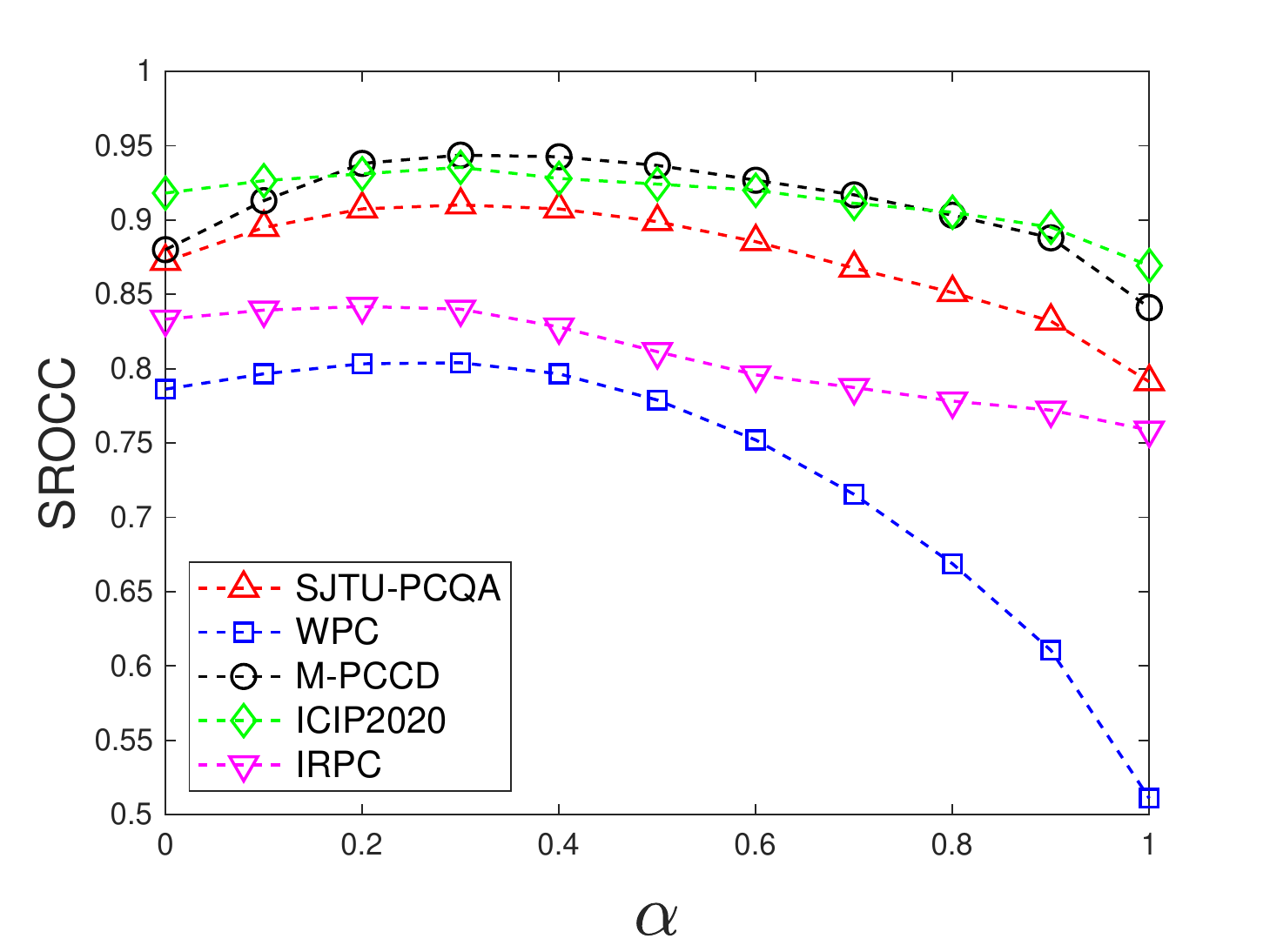}
    \caption{Performance comparison (in term of SROCC) for different $\alpha$ values.}
    \label{fig:ablation_alpha}
\end{figure}

\znote{$\alpha$ determines the relative importance of $F_1$ and $F_2$ in the final quality index. We test the performance of TCDM under different $\alpha$ values and present the results in Fig. \ref{fig:ablation_alpha}. We can see from the figure that as $\alpha$ increases, the performance on the five databases initially improves and subsequently decreases greatly. All databases achieve the best performance with $\alpha=0.3$, which supports our parameter selection in Section \ref{sec:parameter_setting}.}

\subsubsection{Ablation Study}
\znote{There are several components involved in our model, i.e., the complexity-based feature $F_1$ generated by direct multiplication of $F_1^O$ and $F_1^I$ and prediction-based feature $F_2$. To investigate the individual contribution of these components, we perform ablation studies on all five databases and report the results in Table \ref{tab:ablation_feature_combination}. It can be seen that the proposed TCDM achieves the best performance on SJTU-PCQA and WPC. On the other three databases, \textit{w/o} $F_1^O$ is superior on M-PCCD while \textit{w/o} $F_1^I$ performs the best on ICIP2020 and IRPC. Although dropping some components can provide favorable results in certain instances, the performance may degrade on others (e.g., \textit{w/o} $F_1^O$ on ICIP2020 and IRPC, and \textit{w/o} $F_1^I$ on M-PCCD). In comparison, TCDM shows a consistent correlation across various databases, demonstrating the necessity of employing all the components.}

\begin{table}[pt]
  \centering
  \caption{\MakeUppercase{Ablation study (in term of SROCC) for components}}
    \begin{tabular}{|c|c|c|c|c|c|}
    \hline
     Database     & {SJTU-PCQA} & {WPC} & {M-PCCD} & {ICIP2020} & {IRPC}  \\
     \hline
      TCDM &\bf{0.9102}	&\bf{0.8041}	&{0.9437}	&{0.9354}	&{0.8401}	\\
    \hline
    \textit{w/o} $F_1^O$ &0.9006	&0.7538	&\bf{0.9609}	&0.9148	&0.7987 \\
    \textit{w/o} $F_1^I$ &0.8491	&0.7209	&0.8885	&\bf{0.9374}	&\bf{0.8597} \\
    \textit{w/o} $F_1$  &0.8721	&0.7863	&0.8800	&0.9181	&0.8333	\\
    \textit{w/o} $F_2$   &0.7913	&0.5110	&0.8411	&0.8693	&0.7587 	\\
    \textit{w/o} $F_1^O,F_2$ &0.7508	&0.2647	&0.5932	&0.5921	&0.5368  \\
    \textit{w/o} $F_1^I,F_2$ &0.6104	&0.4891	&0.8390	&0.9145	&0.8181 \\
    \hline

    \end{tabular}%
  \label{tab:ablation_feature_combination}%
\end{table}%

\subsection{Computational Complexity}
\znote{This part examines the computational complexity of our model. In Table \ref{tab:running_time}, we report the average running time per sample of various metrics on ICIP2020, where the point number of the reference samples ranges from 0.3 million to 1.41 million. Experiments are performed on a computer with Intel Xeon(R) Gold 6226R CPU @2.9GHz. Note that PCQM is implemented by C++ while other metrics are realized in Matlab R2022a. For TCDM, we calculate the time cost of two implementations that use random sampling (denoted by TCDM-RS) and farthest point sampling (denoted by TCDM-FPS) respectively to generate seeds. The running time of the key-point sampling stage in some metrics (e.g., GraphSIM and MPED) is not included in the total running time because the sampling methods can easily be replaced by random sampling with little efforts. From Table \ref{tab:running_time}, it is observed that TCDM-RS has a moderate complexity among all approaches, while TCDM-FPS is relatively slow due to the FPS operation. PCQM stands out as the fastest method due to its C++ implementation. MPED and M-p2po are also efficient because of their simple forms. MS-GraphSIM presents long running time, primarily because it utilizes a series of multi-scale representations. }

\znote{The computational complexity of TCDM is partly dominated by the kNN search. There exist three steps that require the kNN search in TCDM, including the space segmentation, the auto-regressive prediction in Eq. \eqref{eq:SA-VAR}, and the calculation of $F_2$ in Eq. \eqref{eq:F2_step1}. Given the reference point cloud with $N$ points and $L$ generating seeds, the space segmentation process finds the nearest seed for each point, which requires a complexity of $O(LN)$. For both the auto-regressive prediction and calculation of $F_2$, TCDM finds the $K$ closest points in patches for each query point, which has a complexity of $O(\sum_{l=1}^L N_l^2)$. Considering that $L\ll N$ in our implementation, the running time of kNN comes mainly from the last two steps. To reduce the time cost, one possible solution is to tackle point clouds after voxelization, by which the neighbor relationship can be directly obtained.}

\begin{table}[pt]
    \centering
    \caption{\MakeUppercase{Comparison of Running time}}
    \begin{tabular}{c c}
        \hline
        PCQA metrics &Running time (s)  \\
        \hline
        M-p2po &  5.71 \\
        M-p2pl &  12.60 \\
        PCQM &  2.74 \\
        GraphSIM &  11.85 \\
        MS-GraphSIM & 19.58 \\
        pointSSIM &  5.86  \\
        MPED &  4.65 \\
        TCDM-RS &  10.25 \\
        TCDM-FPS & 23.29 \\
        \hline
    \end{tabular}  
    \label{tab:running_time}
\end{table}

\section{Conclusion}\label{sec:conclusion}
In this paper, we propose a novel full-reference point cloud quality metric from the perspective
of the transformational complexity. Supposing that the point cloud quality can be described as the complexity of transforming the distorted point cloud back to its reference, we utilize an SA-VAR model to encode multiple channels of the reference point cloud in cases with and without its distorted version. Two complexity terms are pooled into a complexity-based feature, and the prediction terms generated by the SA-VAR are further introduced to achieve more robust quality prediction. The proposed metric, denoted as TCDM, shows a consistent and reliable correlation with subjective evaluation on five point cloud quality assessment databases, presenting noticeable gains over state-of-the-art FR-PCQA metrics. Further experiments have supported model generalization by examining its key modules and parameter values.

\ifCLASSOPTIONcompsoc
  \section*{Acknowledgments}
\else
  \section*{Acknowledgment}
\fi
This paper is supported in part by National Natural Science Foundation of China (61971282, U20A20185). The corresponding author is Yiling Xu (e-mail: yl.xu@sjtu.edu.cn).

\ifCLASSOPTIONcaptionsoff
  \newpage
\fi

\bibliographystyle{IEEEtran}
\bibliography{ref}

\end{document}